\documentclass{article}

\usepackage[main,final]{neurips_2026}

\usepackage[utf8]{inputenc}
\usepackage[T1]{fontenc}
\usepackage{multirow}
\usepackage{makecell}
\usepackage{amssymb} 
\usepackage{hyperref}
\usepackage{url}
\usepackage{booktabs}
\usepackage{amsfonts}
\usepackage{amsmath}
\usepackage{amssymb}
\usepackage{nicefrac}
\usepackage{microtype}
\usepackage{textcomp}
\usepackage{xcolor}
\usepackage{graphicx}
\usepackage{algorithm}
\usepackage{algpseudocode}
\usepackage{subcaption}
\usepackage{natbib}
\usepackage[table]{xcolor}
\usepackage{etoc}
\usepackage[titles]{tocloft}
\definecolor{tocblue}{RGB}{65,105,225}
\renewcommand{\contentsname}{\normalfont\Large\textsc{Contents}}
\definecolor{first}{RGB}{120,190,120}    
\definecolor{second}{RGB}{180,220,180}   
\definecolor{third}{RGB}{210,232,210}    

\title{Locality-Controlled Out-of-Distribution Guidance for Robust Regulatory DNA Sequence Design}

\author{%
  Dabin Jeong\\
  Wellcome Sanger Institute\\
  \texttt{dj16@sanger.ac.uk} \\
  \And
  Marwane Bourdim\\
  Wellcome Sanger Institute\\
  \texttt{mb57@sanger.ac.uk} \\
  \And
  Mo Lotfollahi\\
  Wellcome Sanger Institute\\
  \texttt{ml10@sanger.ac.uk} \\
}

\begin{document}

\maketitle

\begin{abstract}
Designing regulatory DNA sequences for targeted gene-expression control is an important challenge.
Recent approaches frame this task as property-conditioned generation, where diffusion models trained on natural sequences are guided toward high predicted activity through reward-based fine-tuning or sampling-time guidance.
We show that single-reward evaluation can obscure reward-hacking behavior.
By decomposing evaluation into property, novelty, and plausibility, we identify an empirically underexplored region of the evaluation space in which all three objectives are satisfied, but no existing baseline succeeds.
{We postulate that this gap arises because prior methods treat guidance primarily as a matter of magnitude.}
In regulatory DNA, however, sequence grammar is spatially nonuniform: motif positions are sensitive to perturbation, while background regions allow more exploration.
We therefore propose \textsc{LocoGen} by reformulating guidance as a problem of \textbf{locality}.
\textsc{LocoGen} learns an invariant motif mask through sequence-level invariance learning to protect motif positions, while applying OOD damping only to background positions.
It operates at sampling time on a frozen diffusion backbone, without retraining the backbone.
Across HepG2, K562, and SK-N-SH, \textsc{LocoGen} reaches this previously underexplored region of the property--novelty--plausibility space.
Code is available at \url{https://github.com/Lotfollahi-lab/LocoGen}.
\end{abstract}

\etocdepthtag.toc{maintext}
\section{Introduction}
\label{sec:intro}
Diffusion models are powerful generative models used in many domains, including images, language, molecules, and biological sequences. Beyond unconditional generation, recent work increasingly focuses on property-conditioned generation, such as generating regulatory DNA sequences that optimize a target property (e.g. gene expression). These advances make diffusion models useful for \emph{de novo} regulatory DNA design, where candidate sequences can be generated and tested in a lab-in-the-loop setting.

We argue that the generated candidates must satisfy three axes simultaneously: (i) high target \textbf{property}, (ii) \textbf{novelty}, and (iii)~\textbf{plausibility} \textemdash preservation of the natural grammar of regulatory DNA (e.g. motif usage diversity/density). No single axis suffices. High property without plausibility fails to reproduce in the wet lab; high novelty without plausibility drifts off the functional manifold; high property without novelty merely reproduces the training distribution. Only when all three axes are jointly satisfied is a candidate worth its wet-lab budget.

\begin{figure}[t]
    \centering
    \includegraphics[width=1.0 \linewidth]{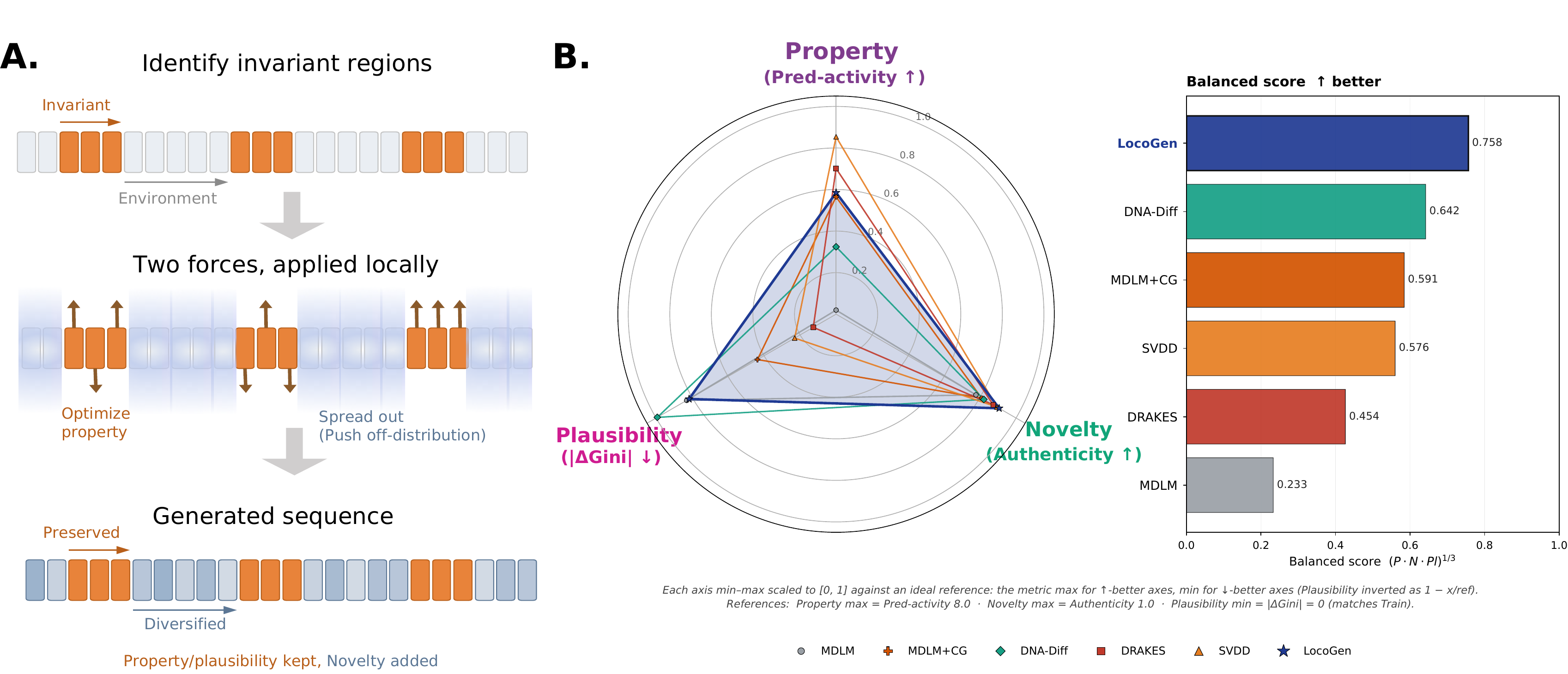}
    \vspace{-10pt}
    \caption{\textsc{LocoGen} balances property, novelty, and plausibility.}
    \label{fig:motivation}
\end{figure}

To date, DNA sequence generative models have largely treated candidate generation as a single-objective problem: maximizing the target property.
However, when evaluation is decomposed into property, novelty, and plausibility, methods that appear strong under single-axis evaluation reveal trade-offs on the remaining axes (Fig.~\ref{fig:motivation}B).
Moreover, their apparent property gains often arise not from targeted regulatory improvement, but from distortions of regulatory motif structure, exposing a common limitation of existing guidance methods (\S\ref{sec:diagnostic}).

Although these methods differ in their reward-hacking mechanisms, they all control optimization through a single \emph{scalar magnitude knob} (e.g. guidance scale) without specifying \emph{where} in the sequence the reward should act.
This is problematic because regulatory DNA grammar is positional: motif sites are highly constrained, whereas intervening background regions are more permissive.
Thus, in the property--novelty--plausibility design space, current baselines fall short of the desired regime: high activity on both readouts, high novelty, and low compositional deviation.
We call this unoccupied region the \textbf{empty corner}, and ask whether guidance can shift the empirical trade-off landscape toward it.

We propose \textsc{LocoGen}, a locality-controlled guidance method for \textit{de novo} regulatory DNA generation (Fig. \ref{fig:motivation}A). Rather than increasing guidance globally, \textsc{LocoGen} controls \emph{where} it acts. A masked separator model identifies foreground and background positions with an invariant mask; during sampling, out-of-distribution (OOD) guidance is applied only to background positions. This generates novel sequences in permissive regions while preserving regulatory motifs, thereby improving novelty without sacrificing enhancer-like composition or functional activity.

\paragraph{Contributions.}
\begin{itemize}
    \item \textbf{Diagnostic framework.} We introduce a three-axis evaluation framework for regulatory DNA generation: property, novelty, and compositional plausibility. It reveals two reward-hacking modes concealed by single-axis evaluation: predictor-specific motif concentration and global compositional drift in previous DNA sequence generative models.

    \item \textbf{Locality-controlled guidance.} We reformulate sampling-time guidance as a locality problem and instantiate it as localized OOD control with an invariant subsequence mask. The method operates on a frozen backbone, requires no retraining, and exposes per-position spatial control absent from standard classifier-guidance methods.

\end{itemize}

\section{Related Work}
\label{sec:related}

\paragraph{Discrete diffusion and flow matching on biological sequences.}
Two frameworks have emerged for generative modeling of categorical sequences. \emph{Discrete diffusion} operates directly on categorical sequences and learns to denoise progressively masked tokens \citep{austin2021structured, sahoo2024simple, lou2024discrete}.
\emph{Continuous diffusion and flow matching} instead work in a continuous relaxation of the categorical space, typically the probability simplex whose vertices correspond to the discrete tokens \citep{stark2024dirichlet, davis2024fisher}
Both have been applied to regulatory DNA design~\citep{stark2024dirichlet,dasilva2026designing,tang2025gumbel}. 
We build on the discrete diffusion line, using a pretrained masked discrete diffusion language model (MDLM) ~\citep{sahoo2024simple} as a backbone and steering generation with inference-time guidance.

\paragraph{Reward-based fine-tuning of diffusion backbones.}
A first family of methods directly fine-tunes the backbone against a reward signal, often framed as reinforcement learning. DRAKES~\citep{wang2024fine} backpropagates a differentiable reward to pre-trained masked discrete diffusion model; DRaFT~\citep{clark2024directly} applies the same differentiable-reward principle in the continuous setting. When the reward is non-differentiable, policy-gradient or gradient-free alternatives are used instead~\citep{cao2025glid2e}.
These methods can each optimize toward a particular property, but they do not expose a post-hoc control knob: changing the target reward generally requires re-specifying the training objective and running a separate fine-tuning pass, even when recent variants reduce its cost or handle multiple objectives jointly.
Although recent methods reduce the cost of retraining~\citep{cao2025glid2e}, they inherently require backbone retraining for each new objective. 

\paragraph{Training-free guidance.}

 Classifier guidance instead allows the guidance signal to be adjusted entirely at inference time through an external scoring network \citep{uehara2025inference, rojas2026improving}. DNA-Diffusion~\citep{dasilva2026designing}, by contrast, conditions on cell-type labels rather than optimizing a reward, and therefore serves as a non-optimizing reference rather than a competing optimizer.
Another method that bypasses backbone retraining is by sampling many candidates and selecting the highest-scoring ones. SVDD~\citep{li2024derivative} performs soft value-based decoding, evaluating generated candidates at each step and resampling proportionally to a learned value. 

\section{Preliminary}
\label{sec:prelim}


\subsection{Notation}
We consider DNA sequences of fixed length $L$ over the nucleotide alphabet
$\mathcal{A} = \{A,T,G,C\}$.
A DNA sequence is denoted by
$X=(x_1,\ldots,x_L)\in\mathcal{A}^L$,
where $x_\ell\in\mathcal{A}$ represents the nucleotide at position $\ell$.
Its one-hot representation is denoted by
$\mathbf{x}\in\{0,1\}^{L\times 4}$,
where each row $\mathbf{x}_\ell\in\{0,1\}^4$
satisfies $\sum_{k=1}^{4}\mathbf{x}_{\ell,k}=1$.

\subsection{Problem formulation}
We consider paired observations $(X, Y)$, where $X \in \mathcal{X} = \mathcal{A}^L$ denotes a DNA sequence and $Y \in \mathcal{Y} \subseteq \mathbb{R}$ denotes an associated scalar property (e.g., target gene activity).
Let $p_\theta(X)$ denote a base generative model over DNA sequences, and let
$p_\phi(Y \mid X)$ denote an auxiliary property predictor.  For a target property value $y^\star$, 
classifier-based guidance aims to reweight the base generative distribution according to the predicted property likelihood as follows,
\begin{equation}
p_{\mathrm{guide}}(X \mid y^\star)
\propto
p_\theta(X)\,p_\phi(y^\star\mid X)^\lambda,
\end{equation}
where $\lambda \geq 0$ controls the guidance strength.
Since our goal is to generate novel regulatory sequences, effective generation guidance must balance three objectives: (i) achieving the target property $y^{\star}$, (ii) producing novel sequences that depart from the training distribution rather than memorizing it, and (iii) maintaining plausibility.

\subsection{Masked Discrete Diffusion Language Models}
\label{sec:MDLM}
MDLM~\citep{sahoo2024simple} generates sequences by
progressively denoising a fully masked input rather
than autoregressively generating tokens.
We extend the alphabet with an absorbing mask
token $\mathtt{[MASK]}$, defining
$\mathcal{A}_{\mathrm{mask}}=\mathcal{A}\cup\{\mathtt{[MASK]}\}$.
Its one-hot representation is denoted by
$\mathbf{m}\in\{0,1\}^5$.
The masked sequence at diffusion time $t$ is denoted by
$\mathbf{z}_t\in\{0,1\}^{L\times5}$.

During training, MDLM learns to recover clean tokens from a masked sequence $\mathbf{z}_t \in \{0,1\}^{L \times 5}$. In the forward process, each clean token in $\mathbf{x}$ is independently replaced with $\mathbf{m}$ with probability $1 - \alpha_t$, yielding $\mathbf{z}_t$, where $\alpha_t$ decreases monotonically from $\alpha_0 \approx 1$ (clean) to $\alpha_1 \approx 0$ (fully masked). 
In the reverse process, a denoising network 
$\boldsymbol{\textbf{x}}_\theta(\mathbf{z}_t,t) : \{0,1\}^{L\times 5} \times [0,1] \to [0,1]^{L\times 5}$
approximates the clean sequence $\mathbf{x}$ by predicting the categorical distribution over clean nucleotides at each position, assigning zero probability to the mask token.

At inference time, generation proceeds by sampling a less-masked state $\mathbf{z}_s$ from the current state $\mathbf{z}_t$, with $s<t$:
\begin{equation}
p_\theta(\mathbf{z}_{s,\ell} \mid \mathbf{z}_t) =
\begin{cases}
\delta_{\mathbf{z}_{t,\ell}}(\mathbf{z}_{s,\ell}),
& \mathbf{z}_{t,\ell} \neq \mathbf{m}, \\[4pt]
\mathrm{Categorical}\!\left(\mathbf{z}_{s,\ell};\,
\dfrac{(1-\alpha_s)\,\mathbf{m} + (\alpha_s - \alpha_t)\,
\boldsymbol{\textbf{x}}_{\theta, \ell}(\mathbf{z}_t, t)}{1 - \alpha_t}
\right),
& \mathbf{z}_{t,\ell} = \mathbf{m}.
\end{cases}
\label{eq:mdlm-reverse}
\end{equation}
Here, $p_\theta(\mathbf{z}_{s,\ell}\mid\mathbf{z}_t)$
denotes the conditional distribution of the token
at position $\ell$ in the next denoising step while $\delta_{\mathbf{z}_{t,\ell}}(\mathbf{z}_{s,\ell})$ denotes a point mass at the current token, ensuring that previously unmasked positions remain unchanged (See \S\ref{app:dirac_delta} for details).

The time $s<t$ corresponds to the next, less-masked diffusion state in the reverse process. Masked positions are either kept as $\mathbf{m}$ or filled in using the model's prediction. Generation terminates when all positions are unmasked.
In this work, we use a pretrained unconditional MDLM over fixed-length DNA sequences as the base model and guide its generation toward specific objectives.

\subsection{Classifier guidance with OOD control}
\label{sec:MOOD}
MOOD~\citep{lee2023exploring} is a score-based diffusion framework that combines OOD control with property guidance to generate novel molecules with desirable properties beyond the training distribution. It operates on continuous representations of molecular graphs. Let $\mathbf{u}_t\in\mathcal{G}$ denote a noise-perturbed molecular graph representation at diffusion time $t$, where $\mathcal{G}$ is the continuous space of molecular graph representations, and let $f_\phi:\mathcal{G}\to\mathbb{R}$ denote a differentiable property scorer trained on molecular data.
In the score-based diffusion framework, classifier guidance modifies the reverse-time SDE drift by adding the gradient of $p_t(y_p\mid\mathbf{u}_t)\propto\exp(\beta_t f_\phi(\mathbf{u}_t))$,
\begin{equation}
d\mathbf{u}_t =
\Bigl[
f_t(\mathbf{u}_t)
-
g_t^2
\Bigl(
\nabla_{\mathbf{u}_t}\log p_t(\mathbf{u}_t)
+
\beta_t\nabla_{\mathbf{u}_t}f_\phi(\mathbf{u}_t)
\Bigr)
\Bigr]dt
+
g_t\,d\bar{w}_t,
\label{eq:classifier_guidance}
\end{equation}
where $p_t$ denotes the marginal density of the forward diffusion process, $f_t$ and $g_t$ are its drift and diffusion coefficients, respectively, $\bar{w}_t$ is a reverse-time Wiener process, and $\beta_t\geq0$ controls the guidance strength. The SDE is integrated backward in time.

MOOD additionally introduces an OOD-control coefficient $\gamma\in[0,1)$ that rescales the generative score:
\begin{equation}
d\mathbf{u}_t =
\Bigl[
f_t(\mathbf{u}_t)
-
(1-\sqrt{\gamma})\,g_t^2
\nabla_{\mathbf{u}_t}\log p_t(\mathbf{u}_t)
-
\beta_t g_t^2
\nabla_{\mathbf{u}_t}f_\phi(\mathbf{u}_t)
\Bigr]dt
+
g_t\,d\bar{w}_t.
\label{eq:mood_uniform}
\end{equation}
where $\gamma=0$ generates samples from property-guided diffusion without OOD control, and increasing $\gamma$ encourages exploration of lower-density regions, thus generating more OOD samples. MOOD applies a single global OOD-control coefficient $\gamma$ to the generative score across all graph components, without adapting this coefficient to their individual contributions to the target property.

\section{Method}
\label{sec:method}
We propose \textsc{Locogen}, a localized sequence generation framework that controls both the magnitude and locality of generated sequences.
The uniform application of guidance and OOD control in Eq.~\ref{eq:mood_uniform} encodes an implicit assumption: that all sequence positions should be perturbed by the same amount. We argue this assumption is not perfectly suited for regulatory DNA. Functional motifs occupy a small fraction of each sequence and are highly constrained by sequence grammar, while the remaining background is comparatively unconstrained. Pushing motifs off-manifold destroys the very signal that makes a sequence functional, while pushing background off-manifold contributes novelty without sacrificing function.

\subsection{Module 1: Masked Separator for Invariant--Environment Decomposition}
\label{sec:masked-separator}

\paragraph{Problem formulation.}
We aim to identify subsequences whose contribution to activity is \emph{invariant} across data {subpopulations}, i.e., positions whose predictive role does not change from one subpopulation to another. A straightforward approach is to train an activity predictor and use saliency, but such scores reflect what a single trained model happens to rely on, without explicitly accounting for invariance. We instead build this requirement into the learning objective: the selected positions are encouraged to predict activity \emph{equally well across subpopulations}. 

\paragraph{Approach.}
Inspired by the invariant-environment decomposition of chemical graphs \citep{li2022learning}, we introduce a learned soft mask $M \in [0,1]^L$ that splits each sequence into an invariant part $x^{\text{inv}}$ and an environment part $x^{\text{env}}$, with $M_\ell$ close to $1$ marking position $\ell$ as invariant and close to $0$
as environment, jointly satisfying: (a) $x^{\text{inv}}$ alone is sufficient to predict activity; (b) its prediction error remains consistent across auto-discovered environments; and (c) $M$ is sparse and contiguous.

\paragraph{Invariant learning for subsequence disentanglement.}
We jointly train a separator $S_\phi$ producing the soft mask $M \in [0,1]^L$ and a predictor $h_\psi$ on the invariant subsequence with disjoint weights. The environment encoder $g_\omega$ on the environment subsequence is randomly initialized and kept frozen. Keeping $g_\omega$ frozen prevents the environment encoder itself from adapting to trivially collapse the environment representations, leaving the separator to determine the partition of sequence information. The environment is still not static since $x^{\text{env}}$ depends on the current mask. 
To prevent information leakage through masked positions, we fill them with a learnable token $b\in\mathbb{R}^K$, where $K=4$ is the nucleotide alphabet size (initialized uniformly to $1/K$):
\begin{equation}
x^{\text{inv}} = x\odot M_K + \bar b\odot(\mathbf{1}{-}M_K), \quad x^{\text{env}} = x\odot(\mathbf{1}{-}M_K) + \bar b\odot M_K,
\label{eq:mix}
\end{equation}
where $M_K=M\otimes\mathbf{1}_K$ and $\bar b=\mathbf{1}_L\otimes b$. The predictor and the environment encoder read out $\hat y^{\text{inv}} = h_\psi(x^{\text{inv}})$ and $h_\psi$'s counterpart $g_\omega(x^{\text{env}})$, respectively.

Since environment labels are unavailable, we infer them via $k$-means clustering with $E{=}3$ clusters on $g_\omega(x^{\text{env}})$, following GIL~\citep{li2022learning}. At the start of each epoch, we freeze a snapshot of $g_\omega$, encode a fixed reference batch, and refit the centroids, which are then cached and used to assign environment labels throughout that epoch. Deriving both the centroids and the embeddings from the same snapshot keeps the label assignment self-consistent. See \S\ref{app:separator-kmeans} for details.

\paragraph{Training objective.}
\begin{equation}
\mathcal{L} = \underbrace{\mathcal{L}_{\text{sta}}}_{\text{Sufficiency}} \;+\; \lambda_{\text{inv}}\,\underbrace{\mathcal{L}_{\text{inv}}}_{\text{Invariance/necessity}} \;+\; 
\lambda_{\text{reg}}\,\mathcal{L}_{\text{reg}} \;+\; 
\lambda_{\text{smooth}}\,\mathcal{L}_{\text{smooth}}.
\label{eq:total-loss}
\end{equation}
Our objective consists of four losses $\mathcal{L}_{\text{sta}}$ for sufficiency, $\mathcal{L}_{\text{inv}}$ for invariant learning via V-REx, $\mathcal{L}_{\text{reg}}$ for density regularization, and $\mathcal{L}_{\text{smooth}}$ for total variation smoothing. $\mathcal{L}_{\text{sta}}$ encourages the predictor to recover $y$ from invariant sequence alone, 
$x^{\text{inv}}$: $\mathcal{L}_{\text{sta}}=\tfrac{1}{B}\sum_n (h_\psi(x^{\text{inv}(n)}) - y^{(n)})^2$, where $n$ indexes the $B$ samples of a minibatch. 
For robust prediction, we adopt variance risk extrapolation (V-REx) \citep{krueger2021out} to encourage consistent predictive performance across inferred environments. Let $\epsilon^{(n)}=(\hat y^{\text{inv}(n)}-y^{(n)})^2$ and $\hat e^{(n)}\in\{1,\dots,E\}$ be the $k$-means environment label of sample $n$. With per-environment risk $R_e = \mathbb{E}_{n:\hat e^{(n)}=e}[\epsilon^{(n)}]$,
\begin{equation}
\mathcal{L}_{\text{inv}} = \mathrm{Var}_e[R_e].
\label{eq:linv}
\end{equation}
Together with $\mathcal{L}_{\text{sta}}$, this encourages the predictor to perform \emph{equally well in every environment}: any environment-specific information leaking into $x^{\text{inv}}$ would inflate the variance and incur a penalty. $\mathcal{L}_{\text{reg}}$ keeps the average mask density near a target $\rho$, and $\mathcal{L}_{\text{smooth}}$ is a fused-lasso total variation penalty \citep{javanmardi2016unsupervised} encouraging contiguous selected blocks. See \S\ref{app:separator-loss} for details on loss definition.
The trained separator and final centroids are saved and remain frozen during sampling.

\subsection{Module 2: Sampler with localized OOD control}
\label{sec:localized_sampling}
\paragraph{Problem.}
The MOOD framework (\S\ref{sec:MOOD}) promotes novel, high-scoring samples by combining sample-level OOD drift with property guidance. For regulatory DNA, this global control is too coarse: without distinguishing constrained motifs from permissive background regions, OOD drift can erase regulatory syntax and compromise activity.
\paragraph{Approach.}
We let damping act only on positions the separator marks as environment. By restricting OOD control to environment positions, we aim to preserve activity-relevant sequence patterns while encouraging exploration in less constrained regions. This is the localization in \textsc{LocoGen}: a structural prior on \emph{where} OOD pressure may act, instead of one global knob trading off the two objectives.
\paragraph{Reverse unmasking process.}
At each step $t \to s$, the reverse process over nucleotide $k$ at position $\ell$ is
\begin{equation}
\tilde p_\theta(x_s \mid x_t)_{\ell,k} \;\propto\;
\underbrace{\exp\!\bigl(d_\ell(\gamma)\cdot\log p_\theta\bigr)}_{\text{(a) localized OOD damping}}
\;\cdot\;
\underbrace{\Delta_{t\to s}}_{\text{(b) MDLM unmask schedule}}
\;\cdot\;
\underbrace{\exp(\eta_t\, \nabla r)}_{\text{(c) property pull}},
\label{eq:locogen-posterior}
\end{equation}
where $\Delta_{t\to s} := (\alpha_s-\alpha_t)/(1-\alpha_t)$ is the probability mass that MDLM assigns to the nucleotide channels at a masked position (Eq.~\ref{eq:mdlm-reverse}), and $\nabla r := \nabla_{\hat x_0} r_\xi(\hat x_0)$ is the gradient of the activity guidance model $r_\xi$ evaluated on the current $x_0$-prediction $\hat x_0$ obtained by softmaxing the nucleotide channels of $\log p_\theta$, following the masked-diffusion parameterization of MDLM.
The three factors push three forces in parallel: (a) flattens background nucleotide distributions toward uniform, (b) preserves MDLM's prescribed unmasking rate, and (c) pulls toward high activity

\paragraph{Where OOD control acts.}
OOD control reduces $\log p_\theta$ by a factor $d_\ell(\gamma) < 1$, flattening the per-position categorical distribution toward uniform and thus injecting randomness. 
We apply OOD damping to positions identified as environment by the separator, using the predicted
mask $M_t=S_\phi(\hat x_0)$:
\begin{equation}
d_{\ell,t}(\gamma) =
\begin{cases}
1-\sqrt{\gamma}, &
M_{t,\ell}<\tau_M \;\text{and}\; t<\tau_{\text{time}},\\
1, & \text{otherwise}.
\end{cases}
\label{eq:dl}
\end{equation}
Restricting damping to masked positions keeps already-decoded nucleotides untouched, consistent with the SUBS parameterisation of MDLM, and is activated only in the latter half of sampling, when $\hat x_0$ provides a more reliable estimate for the separator.
At invariant positions, $d_{\ell,t}(\gamma)=1$,
leaving the base nucleotide distribution unchanged
before property guidance.



\paragraph{Auto-balanced property scale.}
The two forces in Eq.~\eqref{eq:locogen-posterior}, the model prior $\log p_\theta$ and the activity gradient $\nabla r$, live on different scales that drift across reverse steps, so a fixed $\eta_t$ either lets the property pull dominate (yielding unrealistic sequences) or lets it vanish (yielding low-activity sequences). Following \citet{lee2023exploring}, we set $\eta_t$ adaptively at each step so that the gradient magnitude matches the prior up to a user ratio $c$:
\begin{equation}
\eta_t \;=\; \min\!\Bigl(\eta_{\max},\; c \cdot \overline{|\log p_\theta|}_{\text{valid}}\;/\;\overline{|\nabla r|}\Bigr),
\label{eq:eta-balanced}
\end{equation}
where the bars denote means over the four nucleotide channels after excluding the DRAKES sentinel values ($\le -10^5$) at already-unmasked positions, and $\eta_{\max}$ caps the scale when $|\nabla r|$ is small. With $c$ fixed, the property pull stays at a constant fraction of the prior throughout sampling. 

To preserve the MDLM unmasking schedule, we add the scaled gradient only to the four nucleotide log-probabilities and renormalise them so that their total mass remains $\Delta_{t\to s}$; the mask channel is left untouched. Property guidance therefore reshapes the categorical distribution \emph{within} nucleotides without altering when MDLM chooses to unmask a position.

\section{Experiments}
\label{sec:experiments}
\subsection{Experimental Setting}
\label{sec:setup}

\paragraph{Dataset.}
We follow the experimental setup of DRAKES~\citep{wang2024fine} to ensure direct comparability under an established benchmark protocol, adopting the same chromosomal train/test split. The benchmark is based on the publicly available large-scale enhancer dataset of \citet{gosai2024machine}, consisting of approximately 700k 200-bp human enhancer sequences with expression measurements quantified in three human cell lines: HepG2, K562, and SK-N-SH.

\paragraph{Generation protocol and evaluation.}
We follow the DRAKES evaluation scheme~\citep{wang2024fine}: each method generates $N{=}640$ sequences (20 batches $\times$ 32), which are then re-scored by independent held-out oracle to avoid circularity between guidance and evaluation. This oracle has three MPRA readouts (HepG2, K562, SK-N-SH), is never used for any method's guidance or fine-tuning, and serves as our canonical MPRA-property axis. As a second, MPRA-independent functional readout, we additionally report chromatin accessibility from the binary ATAC classifier~\citep{wang2024fine}. For more detailed evaluation protocol, see \S\ref{app:eval}. 

\paragraph{Baselines.}
We compare \textsc{LocoGen} with baselines spanning three categories of regulatory sequence generative models: (1) non-optimizing references: MDLM~\citep{sahoo2024simple} (2) Reward-based fine-tuning: DRAKES~\citep{wang2024fine}, and (3) Training-free guidance: MDLM with classifier guidance (MDLM+CG), SVDD~\citep{li2024derivative}, and DNA-diffusion~\citep{dasilva2026designing}. All baselines are run with their authors' released checkpoints and original decoding hyperparameters. We acknowledge concurrent DNA generative models, e.g GLID2E~\citep{cao2025glid2e}, Ctrl-DNA~\citep{chen2025ctrl}, and GFlowNet~\citep{bengio2021flow}, but do not include them as baselines since no public checkpoints are available and retraining their upstream models is out of scope.
Computation resource requirements are reported in \S\ref{app:computational_cost}.


\subsection{Three-axis evaluation framework}
\label{sec:metrics}

We evaluate a generated set $\mathcal{X}^*=\{x^{*(i)}\}_{i=1}^{N}$ against the training set $\mathcal{X}=\{x^{(j)}\}_{j=1}^{M}$ along three axes, (1) property, (2) novelty, and (3) compositional plausibility under a fixed protocol decoupled from each method's training or guidance loop. All metrics are reported as mean $\pm$ standard deviation across three sampling seeds.

\subsubsection{Property axis.}
We assess functional activity using two complementary held-out predictors from \citet{wang2024fine}: regulatory potency, measured by target gene expression, and chromatin accessibility, which is required for transcription factors to bind regulatory sequences and modulate gene expression. See \S\ref{app:eval-oracle} for details.

\paragraph{Pred-activity} 
We use an independently released oracle that maps a sequence to a scalar prediction of gene expression, with separate heads for cell lines $c \in \{\mathrm{HepG2}, \mathrm{K562}, \mathrm{SK\text{-}N\text{-}SH}\}$. For each cell line, we report the median predicted activity over generated samples.

\paragraph{ATAC\_acc.}
We use an independently released classifier that maps each sequence to a predicted probability of chromatin accessibility, formulated as a binary classification task distinguishing accessible from inaccessible sequences. We report the mean predicted accessibility over generated samples.

\subsubsection{Novelty axis.}

\paragraph{Authenticity~\citep{alaa2022faithful}.}
We report authenticity score to measure novelty and explicitly diagnose training-set memorization.
For each generated sequence $x^{*(i)}$, let $x^{(j_i)}$ be its nearest training neighbour under Euclidean distance $d(\cdot,\cdot)$ in 4-mer frequency space, and let $d_{\mathrm{self}}(x^{(j_i)})$ be that neighbour's leave-one-out 1-NN distance within the training set. We report the fraction of samples that lie farther from their nearest neighbour than the neighbour's natural spacing,
\[
    \mathrm{Authenticity} = \frac{1}{N} \sum_{i=1}^{N} \mathbf{1}\!\left[ d(x^{*(i)}, x^{(j_i)}) > d_{\mathrm{self}}(x^{(j_i)}) \right],
\]
with higher values indicating less memorisation.


\paragraph{Rarity~\citep{han2023rarity}.}
For each training sample $x^{(j)}$, let $\mathrm{NN}_k(x^{(j)})$ be its distance to its $k$-th nearest neighbour in 4-mer frequency space, defining a training $k$-NN ball $B_k(x^{(j)})$ of that radius. The per-sample rarity of an on-manifold generated sequence $x^{*(i)}$ is the radius of the smallest ball containing it,
\[
    r(x^{*(i)}) = \min_{\,j\,:\,x^{*(i)}\in B_k(x^{(j)})}\!\mathrm{NN}_k(x^{(j)}),
\]
with off-manifold samples (outside every $B_k$) excluded so that $r$ isolates on-manifold rarity. To remove dependence on feature scale and sample size, we report $r$ in IQR units relative to the leave-one-out training rarity $r^{\mathrm{train}}$.
Thus $\widetilde{r}=0$ at the training median and $\widetilde{r}=1$ at $p_{75}$, with larger values marking samples in rarer training-density regions.

\subsubsection{Compositional plausibility axis.}
We assess preservation of the natural compositional grammar of regulatory DNA using two sub-metrics, reported as absolute deviations from the training set,
$\Delta = |\,\text{method} - \text{train}\,|$.
Smaller deviations indicate better preservation.

\paragraph{$\Delta$Gini.}
Motif-usage diversity is measured by the Gini index $G$ over motif counts from JASPAR scans of generated and training sequences~\citep{rauluseviciute2024jaspar}.
$G=0$ denotes uniform motif usage, while $G\to1$ indicates domination by a small subset of motifs.
For sorted per-motif counts $f=(f_1,\ldots,f_R)$,
\[
    G(f) = \tfrac{1}{R}\!\left(R+1 - 2\,\tfrac{\sum_{r}(R+1-r)\,f_{(r)}}{\sum_{r} f_{(r)}}\right) \in [0,1].
\]
 See \S\ref{app:eval-metric} for more details.
 
\paragraph{$\Delta$hits.} To measure motif density, we compute hits/seq, the mean number of above-threshold motif occurrences per generated sequence, $\mathrm{hits}/\mathrm{seq} = N^{-1}\!\sum_i h_i$, where $h_i$ is the motif hit count for $x^{*(i)}$. Given $\mathrm{hits}/\mathrm{seq}$ of natural sequences, deviation in either direction indicates motif under- or over-insertion.

\subsection{Diagnostic: two structurally distinct hacking modes}

\label{sec:diagnostic}

Among baselines, both SVDD and DRAKES successfully push predicted activity upward, but along distinct non-natural directions. Natural enhancers rely on the combinatorial cooperation of multiple TFs, with motifs distributed at a characteristic density. Beyond previously reported metrics, we assess how closely the functional motif composition of generated sequences approaches that of natural regulatory sequences by measuring motif usage diversity and density via $\Delta$Gini and $\Delta$hits. A large $\Delta$Gini indicates that generated sequences are dominated by a few repeated motifs, while a large $\Delta$hits indicates stuffed motif within a sequence. In either case, predictor scores can remain high even as the sequence drifts away from natural enhancer composition.

\begin{figure}[h]
    \centering
    \includegraphics[width=0.9\linewidth]{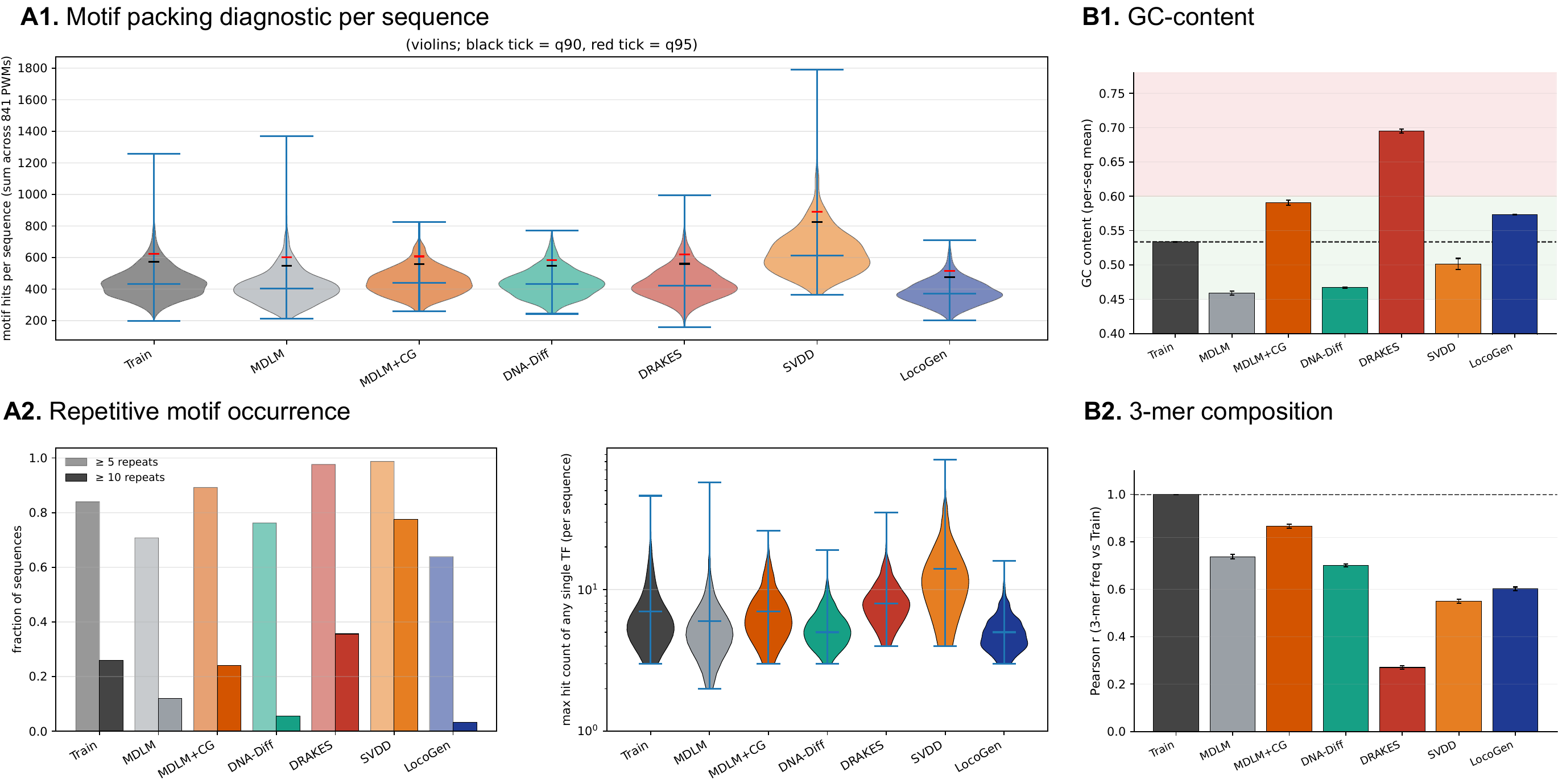}
    \caption{Diagnostics of existing sequence design baselines.}
    \label{fig:diagnostic}
\end{figure}

\paragraph{SVDD: motif inflation.}
In SVDD, motif density per sequence inflates to $1.44\times$ the natural baseline ($\Delta\,\text{hits} = 0.248$, from $0.561$ to $0.809$; Tab.~\ref{tab:bench_hepg2}, Fig.~\ref{fig:diagnostic}A1), while motif usage concentrates on a narrow subset of the JASPAR repertoire ($\Delta\,\text{Gini} = 0.231$, from $0.308$ to $0.539$; Tab.~\ref{tab:bench_hepg2},Fig.~\ref{fig:diagnostic}A2). Repeated use of functionally important motifs is itself biologically plausible; the pathology of SVDD, rather, is that it repeats a single motif up to saturation within a single sequence. Together, these observations suggest that SVDD raises predicted activity by repeatedly stuffing a small set of presumably activity-driving motifs into sequence, rather than by employing the diverse motif vocabulary characteristic of natural enhancers.

\paragraph{DRAKES: GC-rich composition drift.}
Unlike SVDD, DRAKES does not overpack motifs ($\Delta\,\text{hits} = 0.06$; Tab.~\ref{tab:bench_hepg2}, Fig.~\ref{fig:diagnostic}A1-2), but its motif usage is still skewed toward a narrow subset of the JASPAR repertoire ($\Delta\,\text{Gini} = 0.261$; Tab.~\ref{tab:bench_hepg2}). To understand which motifs DRAKES preferentially favors, we examined the over-represented entries and found that they are predominantly GC-rich (Fig. \ref{fig:diagnostic}B1). 
The composition of the generated sequences depart substantially from the natural compositional baseline (Fig.~\ref{fig:diagnostic}B2). Together, these observations suggest that DRAKES inflates predicted activity by drifting toward a generic GC-rich background, rather than by concentrating on the functionally relevant motifs of the target cell line.

\subsection{Reaching the empty corner with \textsc{LocoGen}}
\label{sec:main}

\begin{table}[h]
\centering
\caption{Three axis benchmark in HepG2 cell line. \colorbox{first}{1st}, \colorbox{second}{2nd}, and \colorbox{third}{3rd} best results are highlighted.}
\label{tab:bench_hepg2}
\setlength{\tabcolsep}{2pt}\renewcommand{\arraystretch}{1.10}\small
\begin{tabular}{l c c c c c c}
\toprule
& \multicolumn{2}{c}{Property} & \multicolumn{2}{c}{Novelty} & \multicolumn{2}{c}{Plausibility} \\
\cmidrule(lr){2-3} \cmidrule(lr){4-5} \cmidrule(lr){6-7}
Method & Pred-activity $\uparrow$ & ATAC\_acc\,$\uparrow$ & Authenticity\,$\uparrow$ & Rarity (IQR)\,$\uparrow$ & $|\Delta\,\text{Gini}|\,\downarrow$ & $|\Delta\,\text{hits}|\,\downarrow$ \\
\midrule
Train & 2.406 & 0.244 & 0.728 & - & - & - \\
\midrule
MDLM
& 0.157\,$\pm$\,0.024
& 0.329\,$\pm$\,0.198
& 0.777\,$\pm$\,0.033
& $-0.15$\,$\pm$\,0.03
& \cellcolor{second}0.051\,$\pm$\,0.006
& 0.071\,$\pm$\,0.001 \\
MDLM+CG
& 4.516\,$\pm$\,0.151
& 0.481\,$\pm$\,0.022
& 0.809\,$\pm$\,0.014
& $+0.21$\,$\pm$\,0.02
& 0.169\,$\pm$\,0.013
& \cellcolor{second}0.043\,$\pm$\,0.007 \\
DNA-Diff
& 2.595\,$\pm$\,0.087
& \cellcolor{third}0.688\,$\pm$\,0.293
& 0.821\,$\pm$\,0.015
& $-0.07$\,$\pm$\,0.03
& \cellcolor{first}0.002\,$\pm$\,0.002
& \cellcolor{first}0.010\,$\pm$\,0.002 \\
DRAKES
& \cellcolor{second}5.610\,$\pm$\,0.070
& \cellcolor{first}0.804\,$\pm$\,0.159
& \cellcolor{third}0.873\,$\pm$\,0.004
& \cellcolor{first}$+1.08$\,$\pm$\,0.09
& 0.261\,$\pm$\,0.007
& 0.060\,$\pm$\,0.010 \\
SVDD
& \cellcolor{first}6.830\,$\pm$\,0.024
& 0.442\,$\pm$\,0.375
& \cellcolor{second}0.894\,$\pm$\,0.008
& \cellcolor{second}$+0.84$\,$\pm$\,0.03
& 0.231\,$\pm$\,0.010
& 0.248\,$\pm$\,0.008 \\
\midrule
LocoGen
& \cellcolor{third}4.679\,$\pm$\,0.021
& \cellcolor{second}0.716\,$\pm$\,0.019
& \cellcolor{first}0.906\,$\pm$\,0.006
& \cellcolor{third}$+0.71$\,$\pm$\,0.00
& \cellcolor{third}0.055\,$\pm$\,0.005
& \cellcolor{third}0.057\,$\pm$\,0.003 \\
\bottomrule
\end{tabular}
\vspace{2pt}
\begin{minipage}{0.98\linewidth}
\raggedleft
\scriptsize
\textit{Note.} For natural sequences (Train), $\Delta\,\text{Gini} = 0.308$ and $\Delta\,\text{hits} = 0.561$.
\end{minipage}
\end{table}

Tables~\ref{tab:bench_hepg2}, \ref{tab:bench_k562}, \ref{tab:bench_sknsh}, and Figure~\ref{fig:motivation} show that \textsc{LocoGen} occupies a distinct region of the three-axis benchmark space. Single-axis optimization leads to clear trade-offs: SVDD and DRAKES obtain the strongest property scores but incur large plausibility shifts, whereas DNA-Diff closely matches the training distribution but yields weaker property optimization and limited rarity. \textsc{LocoGen} avoids both failure modes. It achieves competitive property optimization, the best authenticity, positive rarity, and top-three performance on both plausibility metrics. 
Thus, our claim is not that \textsc{LocoGen} is absolutely superior on every metric, but that it is the property-optimizing method that fills the empty corner of the design space: maintaining acceptable performance on property, novelty, and plausibility simultaneously.
We further verify that the masked separator learns a meaningful invariant/environmental decomposition rather than an arbitrary partition.
The invariant subsequence remains close to the natural baseline, is enriched for activity-associated JASPAR motifs, and has a larger functional impact under ablation than random regions; see \S\ref{app:effect_of_masking} for details.


\subsubsection{Ablation study}
\label{sec:ablation}
To better understand what drives \textsc{LocoGen}'s performance, we investigate two complementary questions: (i) Which components of \textsc{LocoGen} contribute to its performance? (ii) Can a non-generative approach based on localization and shuffling achieve comparable results?

\paragraph{(i)}Table~\ref{tab:locality_ablation} disentangles the contributions of guidance and locality in \textsc{LocoGen}. \textit{CG} applies classifier guidance alone ($\gamma=0$), while \textit{Loc.~OOD} (\textit{Learnt}) represents the full model, restricting OOD damping to the mask separator-identified background. \textit{Glob.~OOD} control positions at random rather than from the masked separator signal with minimal overlap with the invariant region (Fig.~\ref{fig:learnt_vs_random}). In contrast, \textit{Attack} reverses property guidance at invariant positions, actively disrupting the regions identified by the masked separator.
Disabling \textit{CG} substantially reduces both property optimization and novelty while leaving plausibility largely unaffected, highlighting the critical role of classifier guidance in property optimization (rows 1-3 vs. the last row).
With \textit{CG} enabled, replacing \textit{Loc.~OOD} (\textit{Learnt}) with \textit{Glob.~OOD} mildly but consistently degrades property and novelty metrics (row 5 vs. the last row), providing additional evidence that the separator identifies activity-sensitive positions. Meanwhile, the comparison against global OOD damping supports a modest but consistent benefit from localizing OOD control.  Removing OOD control substantially degrades all metrics except $|\Delta\text{hits}|$ (row 4 vs. the last row). Thus, classifier guidance with OOD control helps discover sequences with stronger properties and greater novelty, while concentrating motif usage on a narrower subset of motifs; the total number of motif hits remains close to the reference distribution. 
We further examine the effect of OOD control strength $\gamma$ to assess how strongly to apply distributional calibration during sampling. See Appendix~\S\ref{app:lambda_sweep} for the full results.

\begin{table}[h]
\centering
\caption{Locality ablation on HepG2 cell line. }
\label{tab:locality_ablation}
\setlength{\tabcolsep}{3pt}\renewcommand{\arraystretch}{1.15}\footnotesize
\begin{tabular}{c c l c c c c c c}
\toprule
\multicolumn{3}{c}{Component} & \multicolumn{2}{c}{Property} & \multicolumn{2}{c}{Novelty} & \multicolumn{2}{c}{Plausibility} \\
\cmidrule(lr){1-3} \cmidrule(lr){4-5} \cmidrule(lr){6-7} \cmidrule(lr){8-9}
CG & \shortstack{Glob. \\OOD} & \multicolumn{1}{c}{\shortstack{Loc.\\OOD}} & Pred-activity $\uparrow$ & ATAC\_acc\,$\uparrow$ & Authenticity\,$\uparrow$ & Rarity (IQR)\,$\uparrow$ & $|\Delta\,\text{Gini}|\,\downarrow$ & $|\Delta\,\text{hits}|\,\downarrow$ \\
\midrule
& & \multicolumn{1}{c}{--}
& 0.157\,$\pm$\,0.024 & 0.329\,$\pm$\,0.198 & 0.777\,$\pm$\,0.033
& $-0.15$\,$\pm$\,0.03 & 0.051\,$\pm$\,0.006 & 0.071\,$\pm$\,0.001 \\
& \checkmark & \multicolumn{1}{c}{--}
& 0.300\,$\pm$\,0.028 & 0.314\,$\pm$\,0.008 & 0.946\,$\pm$\,0.002
& $+0.46$\,$\pm$\,0.03 & 0.246\,$\pm$\,0.002 & 0.167\,$\pm$\,0.007 \\
& & Learnt
& 0.309\,$\pm$\,0.020 & 0.311\,$\pm$\,0.009 & 0.947\,$\pm$\,0.003
& $+0.46$\,$\pm$\,0.03 & 0.244\,$\pm$\,0.004 & 0.166\,$\pm$\,0.006 \\

\checkmark & & \multicolumn{1}{c}{--}
& 4.516\,$\pm$\,0.151 & 0.481\,$\pm$\,0.022 & 0.809\,$\pm$\,0.014
& $+0.21$\,$\pm$\,0.02 & 0.169\,$\pm$\,0.013 & 0.043\,$\pm$\,0.007 \\
\checkmark & \checkmark & \multicolumn{1}{c}{--}
& 4.608\,$\pm$\,0.061 & 0.691\,$\pm$\,0.041 & 0.887\,$\pm$\,0.012
& $+0.62$\,$\pm$\,0.11 & 0.065\,$\pm$\,0.021 & 0.044\,$\pm$\,0.015 \\
\checkmark & & Attack
& 3.881\,$\pm$\,0.038 & 0.656\,$\pm$\,0.004 & 0.887\,$\pm$\,0.012
& $+0.64$\,$\pm$\,0.06 & 0.025\,$\pm$\,0.007 & 0.083\,$\pm$\,0.006 \\
\midrule
\checkmark & & Learnt
& 4.679\,$\pm$\,0.021 & 0.716\,$\pm$\,0.019 & 0.906\,$\pm$\,0.006
& $+0.71$\,$\pm$\,0.00 & 0.055\,$\pm$\,0.005 & 0.057\,$\pm$\,0.003 \\
\bottomrule
\end{tabular}
\end{table}

\paragraph{(ii)}To distinguish the contribution of learned localization from that of generative optimization, we conducted a non-generative control experiment based on a segment-and-shuffle strategy (Tab. \ref{tab:segment_shuffle}). See details in \S\ref{app:seg_and_shuffle}. 
LearnedMask outperformed heuristic segmentation methods (e.g., predicted activity of 0.442 vs. 0.359 for GC-content masking under uniform shuffling), suggesting that learned localization better identifies activity-relevant regions. Nevertheless, preserving these regions while shuffling the background remained substantially less effective than full \textsc{LocoGen} (predicted activity: 0.442 vs. 4.679; ATAC accessibility: 0.280 vs. 0.716). These results suggest that localization alone is insufficient to reproduce LocoGen's performance, highlighting the importance of combining learned localization with a property-guided generative model.

\subsection{Robustness to evaluation configuration}
To verify that the balanced three-axis profile is not an artifact of our evaluation setup, we evaluated \textsc{LocoGen} under (1) independent oracles and (2) a hyperparameter sweep.
We first re-evaluated all methods with two publicly available alternative oracles released by prior work~\citep{li2024derivative, genentech_human_atac_catlas_model}. See
\S\ref{app:robust_oracle} for details. Swapping in these oracles leaves most of the ranking intact (Tab. \ref{tab:robustness-hepg2},\ref{tab:robustness-k562},\ref{tab:robustness-sknsh}).
Its three-axis profile is therefore not tied to a specific oracle.
Next, we swept \textsc{LocoGen} and every baseline over the
ranges given in each method's paper or official repository; see \S\ref{app:robust_hyperparam} for
details. Across these sweeps, \textsc{LocoGen} retains its balanced three-axis
advantage, confirming that the result does not rely on a cherry-picked hyperparameter configuration.

\section{Conclusion}
We studied sampling-time guidance for regulatory DNA generation and showed that single-axis evaluation can obscure distinct reward-hacking modes.
By decomposing evaluation into property, novelty, and plausibility, we identified a desirable region that existing baselines fail to reach.
We showed that this gap can be addressed by treating guidance as a problem of \textbf{locality} (which positions to modify) rather than scalar strength.
Implemented as localized OOD control with an invariant motif mask, this approach improves the three-axis balance on a frozen backbone without retraining.
More broadly, our results suggest that locality-aware guidance may be useful in other structured domains with spatially or functionally nonuniform grammars.
See \S\ref{app:discussion} for further discussions on limitations and impact of this work.

\begin{ack}
We acknowledge core funding from the Wellcome Trust (WT220540/Z/20/A), as well as funding and support from Open Targets.
\end{ack}

\newpage
\bibliographystyle{plainnat}
\bibliography{refs}

\newpage

\appendix
\etocdepthtag.toc{appendix}
\section*{Supplementary material}

\etocsettagdepth{maintext}{none}
\etocsettagdepth{appendix}{subsection}
\etocstandardlines
\etocclasstocstyle
\etocsettocstyle{\section*{\contentsname}}{}
\tableofcontents

\clearpage

\setcounter{figure}{0}
\setcounter{table}{0}
\setcounter{equation}{0}
\renewcommand{\thefigure}{S\arabic{figure}}
\renewcommand{\thetable}{S\arabic{table}}
\renewcommand{\theequation}{S\arabic{equation}}
\renewcommand{\thesection}{S\arabic{section}}
\renewcommand{\theHfigure}{S\arabic{figure}}
\renewcommand{\theHtable}{S\arabic{table}}
\renewcommand{\theHequation}{S\arabic{equation}}

\section{Supplementary methods}
\label{app:method}

This section fixes the notation used throughout the paper
(\S\ref{app:notation}, Table~\ref{tab:notation}),
gives the derivation details of the MDLM reverse step omitted from
\S\ref{sec:MDLM} (\S\ref{app:mdlm}), and then documents the two
\textsc{LocoGen} modules in the order in which they are introduced in
\S\ref{sec:method}: the masked separator (\S\ref{app:separator}) and the
localization-guided sampler (\S\ref{app:sampler}).

\subsection{Notation}
\label{app:notation}

\begin{table}[h!]
\centering
\caption{Summary of notation. Symbols are grouped by the section in which they are introduced.}
\label{tab:notation}
\small
\begin{tabular}{ll}
\toprule
\textbf{Symbol} & \textbf{Description} \\
\midrule
\multicolumn{2}{l}{\emph{Sequences and diffusion (\S\ref{sec:prelim})}} \\
$\mathcal{A}$ & Nucleotide alphabet $\{A,T,G,C\}$ \\
$K=4$ & Alphabet size \\
$L$ & DNA sequence length ($L=200$ in all experiments) \\
$\ell,\,k$ & Position index and nucleotide-channel index \\
$n,\,B$ & Sample index within a minibatch and minibatch size \\
$X\in\mathcal{A}^L$ & Discrete DNA sequence \\
$x_\ell\in\mathcal{A}$ & Nucleotide at position $\ell$ \\
$\mathbf{x}\in\{0,1\}^{L\times K}$ & One-hot encoded DNA sequence; $\mathbf{x}_\ell$ its row at position $\ell$ \\
$\mathtt{[MASK]},\ \mathbf{m}$ & Mask symbol and its one-hot representation in $\{0,1\}^{K+1}$ \\
$\mathbf{z}_t\in\{0,1\}^{L\times(K+1)}$ & Masked sequence (diffusion state) at diffusion time $t$ \\
$s<t$ & Next, less-masked diffusion time in the reverse process \\
$\alpha_t$ & MDLM masking schedule \\
$\Delta_{t\to s}$ & Unmasking mass $(\alpha_s-\alpha_t)/(1-\alpha_t)$ (Eq.~\ref{eq:mdlm-reverse}) \\
$p_\theta$ & Frozen base (MDLM) generative model \\
$p_\phi(Y\mid X)$ & Auxiliary property predictor \\
$y^\star$ & Target property value \\
$\lambda$ & Property guidance strength in the guided distribution \\
$\mathbf{u}_t,\ f_\phi,\ \beta_t$ & MOOD diffusion state, property scorer, and guidance strength \\
$\gamma\in[0,1)$ & OOD-control strength \\
\addlinespace
\multicolumn{2}{l}{\emph{Masked separator (\S\ref{sec:masked-separator})}} \\
$S_\phi$ & Separator producing the soft mask \\
$h_\psi,\ g_\omega$ & Invariant predictor and frozen environment encoder \\
$M\in[0,1]^L$ & Soft invariance mask; $M_K=M\otimes\mathbf{1}_K$ its channel-broadcast form \\
$x^{\text{inv}},\ x^{\text{env}}$ & Invariant and environment subsequences (Eq.~\ref{eq:mix}) \\
$b\in\mathbb{R}^K,\ \bar b$ & Learnable fill token and its sequence-broadcast form \\
$E=3$ & Number of auto-discovered environments \\
$\hat e^{(n)},\ R_e,\ c_e$ & Environment label, per-environment risk, and centroid \\
$\rho$ & Target mask density \\
$\lambda_{\text{inv}},\lambda_{\text{reg}},\lambda_{\text{smooth}}$ & Loss weights in Eq.~\ref{eq:total-loss} \\
\addlinespace
\multicolumn{2}{l}{\emph{Localization-guided sampler (\S\ref{sec:localized_sampling})}} \\
$r_\xi,\ \nabla r$ & Activity guidance model and its gradient w.r.t.\ $\hat x_0$ \\
$\hat x_0$ & Current $x_0$-prediction obtained by softmaxing the nucleotide channels \\
$M_t=S_\phi(\hat x_0)$ & Mask predicted at diffusion time $t$ \\
$d_{\ell,t}(\gamma)$ & Localized OOD damping factor (Eq.~\ref{eq:dl}) \\
$\tau_M,\ \tau_{\text{time}}$ & Mask and diffusion-time thresholds gating the damping \\
$\eta_t,\ \eta_{\max},\ c$ & Property-pull scale, its cap, and the target magnitude ratio (Eq.~\ref{eq:eta-balanced}) \\
\bottomrule
\end{tabular}
\end{table}

\subsection{Reverse process of MDLM}
\label{app:mdlm}
\subsubsection{Factorization over positions}
The per-position transition in Eq.~\ref{eq:mdlm-reverse} is applied
independently at every position, so the reverse transition of the whole
sequence factorizes as
\begin{equation}
p_\theta(\mathbf{z}_s\mid\mathbf{z}_t)
=
\prod_{\ell=1}^{L}
p_\theta(\mathbf{z}_{s,\ell}\mid\mathbf{z}_t).
\end{equation}

\subsubsection{Point mass at already-decoded positions}
\label{app:dirac_delta}
At diffusion time $t$, let $a_{t,\ell}\in\mathcal{A}_{\mathrm{mask}}$
denote the categorical token at position $\ell$, with one-hot
representation
\begin{equation}
\mathbf{z}_{t,\ell}
=\operatorname{onehot}(a_{t,\ell})
\in\{0,1\}^{5},
\qquad
\sum_{k=1}^{5}[\mathbf{z}_{t,\ell}]_k=1 .
\end{equation}
The first branch of Eq.~\ref{eq:mdlm-reverse} is then the point mass
\begin{equation}
\delta_{\mathbf{z}_{t,\ell}}(\mathbf{z}_{s,\ell}) =
\begin{cases}
1, & \mathbf{z}_{s,\ell} = \mathbf{z}_{t,\ell},\\
0, & \text{otherwise},
\end{cases}
\end{equation}
which freezes every position whose token has already been decoded; only
positions still carrying $\mathbf{m}$ are resampled. This is the SUBS
parameterization that \textsc{LocoGen} inherits, and it is the reason the
damping in \S\ref{app:sampler} can act only on still-masked positions.

\subsection{Masked separator}
\label{app:separator}
\subsubsection{Backbone architecture}
\label{app:separator-arch}
Each of the three networks $S_\phi$, $h_\psi$, and $g_\omega$ is a stack of
dilated 1D convolutions of identical capacity to the DRAKES classifier
\citep{wang2024fine}, with the hyperparameters of
Table~\ref{tab:hparams-separator}: 5 stacks of 5 dilations with rates
$\{1,1,4,16,64\}$, kernel size 9, hidden width $H=128$, residual
connections within each stack, and ReLU activations. The input is the
one-hot tensor $\mathbf{x}\in\{0,1\}^{L\times K}$ with $L=200$ and $K=4$.
The separator head applies a position-wise sigmoid to produce
$M\in[0,1]^L$, with its output bias initialized at $\mathrm{logit}(\rho)$
so that $\mathbb{E}[M]\approx\rho$ at the start of training.
The predictor reads out the invariant subsequence through a 2-layer head
on global mean pooling,
\begin{equation}
\hat y^{\text{inv}} \;=\; W_2\,\mathrm{ReLU}\!\Bigl(W_1\,\tfrac{1}{L}\textstyle\sum_{\ell=1}^{L} h_\psi(x^{\text{inv}})_\ell\Bigr),\qquad W_1 \in \mathbb{R}^{H \times H},\; W_2 \in \mathbb{R}^{1 \times H},
\end{equation}
while the environment encoder produces a single embedding per sequence,
\begin{equation}
g_\omega(x^{\text{env}}) \;=\; \tfrac{1}{L}\textstyle\sum_{\ell=1}^{L} \tilde g_\omega(x^{\text{env}})_\ell \;\in\; \mathbb{R}^H ,
\label{eq:env-embedding}
\end{equation}
where $\tilde g_\omega$ denotes the convolutional trunk before pooling. We
write $g_\omega(x^{\text{env}})$ for the pooled embedding throughout, as in
\S\ref{sec:masked-separator}.

The three networks share architecture but \emph{not} weights, and only
$S_\phi$ and $h_\psi$ are trained: $g_\omega$ is randomly initialized and
kept frozen. A trainable environment encoder can lower
$\mathcal{L}_{\text{inv}}$ by collapsing the embedding space itself rather
than by improving the partition; tying $g_\omega$ to $h_\psi$ has the same
effect, since the predictor's collapse attractor
($x^{\text{inv}}\!\to\!\bar b$, $\hat y^{\text{inv}}$ constant) would then
corrupt the embedding space used for $k$-means
(\S\ref{app:separator-kmeans}). Freezing $g_\omega$ leaves the separator as
the only route by which the environment partition can change, which is what
makes the V-REx penalty informative about the mask.

\subsubsection{Auxiliary loss terms}
\label{app:separator-loss}
\paragraph{Mask anchoring ($\mathcal{L}_{\text{reg}}$).}
A symmetric quadratic penalty pulling the average mask density to the
target $\rho$, with an optional bimodal sub-term:
\begin{equation}
\mathcal{L}_{\text{reg}} \;=\; (\bar M - \rho)^2 \;+\; \lambda_{\text{bim}}\,\overline{M(1-M)},\qquad \bar M \;=\; \tfrac{1}{BL}\textstyle\sum_{n,\ell} M^{(n)}_\ell .
\end{equation}
The first term is expected-$L_0$ anchoring \citep{louizos2017learning}. We
set $\lambda_{\text{bim}}\!=\!0$: with $\lambda_{\text{bim}}\!>\!0$ the
$\{0,1\}$ valleys become attractors and combine with the mean term to
collapse the separator within the first epoch. The target density $\rho$ is
held as a \emph{frozen buffer} rather than a trainable parameter; when made
learnable it gets pulled toward $\bar M$, which hollows out the penalty
whenever the mask collapses.
\paragraph{Spatial smoothness ($\mathcal{L}_{\text{smooth}}$).}
A fused-lasso total-variation penalty on adjacent mask values
\citep{javanmardi2016unsupervised}:
\begin{equation}
\mathcal{L}_{\text{smooth}} \;=\; \tfrac{1}{B(L-1)}\textstyle\sum_{n}\sum_{\ell=1}^{L-1}\bigl|M^{(n)}_{\ell+1} - M^{(n)}_\ell\bigr| .
\end{equation}
This encourages contiguous selected blocks rather than scattered single
positions, consistent with the cis-regulatory grammar in which functional
elements occupy short contiguous motifs.
\paragraph{V-REx variance estimator.}
$\mathcal{L}_{\text{inv}}$ in Eq.~\ref{eq:linv} is computed in its biased
form. Environments containing fewer than two samples are dropped from the
variance estimate; if fewer than two environments survive, we set
$\mathcal{L}_{\text{inv}}\!=\!0$ for that batch.

\subsubsection{Environment inference by $k$-means}
\label{app:separator-kmeans}
Because $x^{\text{env}}$ is defined by the current mask, the inferred
environments shift as $S_\phi$ is updated. We therefore refit $k$-means at
the start of every epoch, so that the V-REx penalty is always measured
against the partition induced by the current mask, and hold the labels
fixed within the epoch, which prevents the separator from lowering
$\mathrm{Var}_e[R_e]$ by reshuffling the environments themselves (the
standard GIL alternating optimization \citep{li2022learning}). Writing $j$
for the epoch index, one round proceeds as follows.
\paragraph{Step 1 --- Snapshot collection.}
At the start of each epoch $j \geq t_{\text{warmup}}$, a copy $\Theta_j$ of
the current parameters is frozen; since $g_\omega$ is already frozen, the
snapshot pins the separator, and hence the mask that defines
$x^{\text{env}}$. Up to $N_{\text{km}}\!=\!8192$ samples are drawn from the
train loader and their environment embeddings
$g_\omega(x^{\text{env}};\Theta_j)$ are collected. For this pass alone we
use \emph{hard-input} mixing: the fill token $b$ is \textbf{not} injected,
so $k$-means clusters on actual background nucleotide content rather than
on a $b$-contaminated mixture.
\paragraph{Step 2 --- $k$-means fit.}
Lloyd's algorithm \citep{lloyd1982least} is run with $E\!=\!3$ clusters and
$T_{\text{km}}\!=\!20$ iterations, initialized by random-sample seeding
indexed by the epoch number. The resulting centroids
$\{c_e\}_{e=1}^{E}\subset\mathbb{R}^H$ are cached.
\paragraph{Step 3 --- Minibatch assignment.}
During each training step within the epoch, samples are assigned to the
nearest cached centroid under squared $\ell_2$ distance:
\begin{equation}
\hat e^{(n)} \;=\; \arg\min_{e\in\{1,\dots,E\}} \bigl\| g_\omega(x^{\text{env}(n)};\Theta_j) - c_e \bigr\|_2^2 .
\end{equation}
These labels feed $\mathcal{L}_{\text{inv}}$ (Eq.~\ref{eq:linv}). Because
both the centroids and the embeddings are taken from the same frozen
snapshot $\Theta_j$, environment labels and centroids stay mutually
consistent for the whole epoch.
\paragraph{Warmup.}
For the first $t_{\text{warmup}}\!=\!1$ epoch no centroids exist; we set
$\mathcal{L}_{\text{inv}}\!=\!0$ and train as pure ERM on $x^{\text{inv}}$.
Subsequently the refit runs at the start of every epoch.
\paragraph{Persistence.}
The final centroids $\{c_e\}$ are saved with the model checkpoint for
reproducibility and for the environment diagnostics of
\S\ref{app:effect_of_masking}. The sampler of \S\ref{app:sampler} consumes
only the separator output $M_t=S_\phi(\hat x_0)$ and does not consult the
centroids.

\begin{table}[h!]
\centering
\caption{Hyperparameters for the masked separator (\S\ref{app:separator}).}
\label{tab:hparams-separator}
\small
\begin{tabular}{llll}
\toprule
Symbol & Value & Description & Source \\
\midrule
$H$                       & 128              & hidden dim of CNN backbones      & \texttt{model.hidden\_dim} \\
stacks                    & 5                & dilated-CNN stack depth          & \texttt{model.num\_cnn\_stacks} \\
dropout                   & 0.1              & per-conv dropout                 & \texttt{model.dropout} \\
$\rho$                    & 0.3              & target invariant mask density    & \texttt{init\_rho} \\
$\lambda_{\text{inv}}$    & 1.0              & V-REx weight                     & config \\
$\lambda_{\text{reg}}$    & 5.0              & mask-anchoring weight            & config \\
$\lambda_{\text{smooth}}$ & 0.1              & fused-lasso TV weight            & config \\
$\lambda_{\text{bim}}$    & 0.0              & bimodal sub-term weight (disabled) & config \\
$E$                       & 3                & number of $k$-means environments & \texttt{num\_envs} \\
$T_{\text{km}}$           & 20               & $k$-means Lloyd iterations        & \texttt{kmeans\_iters} \\
$N_{\text{km}}$           & 8192             & $k$-means subsample cap           & \texttt{kmeans\_samples} \\
$t_{\text{warmup}}$       & 1 epoch          & ERM-only warmup                  & \texttt{warmup\_epochs} \\
optimiser                 & Adam             & lr $10^{-4}$, grad-clip max-norm 1.0 & \texttt{train.lr} \\
\bottomrule
\end{tabular}
\end{table}

\subsection{Localization-guided sampler}
\label{app:sampler}
The sampler operates entirely at inference time on frozen components: the
MDLM prior $p_\theta$, the separator $S_\phi$, and the activity guidance
model $r_\xi$. At each reverse step $t\!\to\!s$ it evaluates $p_\theta$
once, forms the $x_0$-prediction $\hat x_0$, queries the separator to obtain
$M_t=S_\phi(\hat x_0)$, and combines the three factors of
Eq.~\ref{eq:locogen-posterior}: the localized damping
$d_{\ell,t}(\gamma)$ of Eq.~\ref{eq:dl}, the MDLM unmasking mass
$\Delta_{t\to s}$, and the auto-balanced property pull
$\eta_t\nabla r$ of Eq.~\ref{eq:eta-balanced}. Damping is gated by two
thresholds: it acts only where $M_{t,\ell}<\tau_M$, i.e.\ on environment
positions, and only for $t<\tau_{\text{time}}$, i.e.\ in the second half of
sampling, where $\hat x_0$ is informative enough for $S_\phi$ to classify
positions reliably. Because already-decoded positions are frozen by the
point mass of \S\ref{app:dirac_delta}, damping is in effect restricted to
still-masked positions.

\paragraph{Sentinel-aware normalization.}
The auto-balance of Eq.~\ref{eq:eta-balanced} normalizes the property pull
by the mean magnitude of the prior log-probabilities. DRAKES
\citep{wang2024fine} writes a large negative sentinel at positions whose
token has already been determined, in order to block re-sampling. Taken
without filtering, these sentinel entries dominate
$\overline{|\log p_\theta|}$, inflate the ratio, and pin $\eta_t$ at its cap
$\eta_{\max}$ at every step, which silently disables auto-balance. We
therefore restrict the mean to non-sentinel entries
($\log p_\theta>-10^5$). The same correction is applied when running the
classifier-guided baselines, so that no method is penalized by this
implementation detail.

\paragraph{Finalization.}
Tokens are drawn from the renormalized categorical distribution over the
four nucleotide channels, with the copy rule of MDLM keeping already
unmasked positions fixed. After the final reverse step, one additional
forward pass with per-position $\arg\max$ removes any residual mask tokens,
as in DRAKES \citep{wang2024fine}.

Remaining architecture and configuration details are provided in the code;
hyperparameter settings are summarized in Table~\ref{tab:hparams-sampler}.

\begin{table}[h!]
\centering
\caption{Hyperparameters for the localization-guided sampler (\S\ref{app:sampler}).}
\label{tab:hparams-sampler}
\small
\begin{tabular}{lll}
\toprule
Symbol & Value & Description \\
\midrule
$\gamma$          & 0.7     & OOD damping strength ($d_{\ell,t}\!=\!1\!-\!\sqrt\gamma$, Eq.~\ref{eq:dl}) \\
$\tau_M$          & 0.5     & mask threshold; damping acts where $M_{t,\ell}\!<\!\tau_M$ \\
$\tau_{\text{time}}$ & 0.5  & damping time threshold (active only for $t\!<\!\tau_{\text{time}}$) \\
$c$               & 3.0     & target property-pull-to-prior magnitude ratio (Eq.~\ref{eq:eta-balanced}) \\
$\eta_{\max}$     & 20      & cap on $\eta_t$ when $|\nabla r|$ is small \\
sentinel cutoff   & $-10^5$ & threshold below which $\log p_\theta$ is excluded from the norm \\
batch / seeds     & 32 / 3  & sampling batch size and number of seeds \\
$N$               & 640     & sequences per method per cell line per seed ($20\!\times\!32$) \\
\bottomrule
\end{tabular}
\end{table}



\subsection{Computational cost}
\label{app:computational_cost}
All experiments were run on a single NVIDIA H200 GPU per job, with 16 CPU cores and 40 GB of host RAM.
Training the masked separator required approximately 5.5 hours per epoch, with peak host-side RAM usage of 4.6 GB.
For sequence generation, sampling 640 sequences with 128 reverse-diffusion steps took 191 wall-clock seconds per random seed, with peak host-side RAM usage of 3.5 GB.

\section{Evaluation protocol}
\label{app:eval}
This section presents further details omitted from the main text due to space constraints.

\subsection{Evaluation metric}
\label{app:eval-metric}
\paragraph{$\Delta$Gini.} We scan generated/training sequences against the JASPAR motif database TF position-specific scoring matrices~\citep{rauluseviciute2024jaspar}, thresholded at false positive rate $10^{-3}$ under a uniform ACGT background. Following \citet{gosai2024machine}, motifs are restricted to those whose constituent TF subunits are all expressed at $\mathrm{TPM}>1$ in the target cell type. The Gini coefficient over per-motif counts $f=(f_1,\ldots,f_R)$ sorted by ascending order is
\[
    G(f) = \tfrac{1}{R}\!\left(R+1 - 2\,\tfrac{\sum_{r}(R+1-r)\,f_{(r)}}{\sum_{r} f_{(r)}}\right) \in [0,1].
\]

\subsection{Baseline re-scoring and the DRAKES oracle}
\label{app:eval-oracle}
To avoid circularity from using the same predictor for both
classifier-guided generation and evaluation, we used separate independent
oracles for generation guidance and sequence evaluation, following the
DRAKES protocol \citep{wang2024fine}. Specifically, \texttt{reward\_oracle\_ft.ckpt}
is used only for guidance or reward-based fine-tuning, whereas
evaluation is performed using the separately distributed
\texttt{reward\_oracle\_eval.ckpt}. This held-out evaluator is a gReLU
\texttt{EnformerPretrainedModel} with three MPRA readouts corresponding
to HepG2, K562, and SK-N-SH, and never contributes to the guidance or
fine-tuning signal of any method. As a second functional readout
independent of MPRA, we additionally report chromatin accessibility from
the binary ATAC classifier \texttt{binary\_atac\_cell\_lines.ckpt}
\citep{wang2024fine}; the two predictors are trained independently.

\section{Supplementary results}
\label{app:result}


\subsection{Benchmark analysis on other cell lines.}
\begin{table}[h!]
\centering
\caption{Three axis benchmark in K562 cell line ($n{=}640 \times 3$ seeds, mean $\pm$ std). \colorbox{first}{1st}, \colorbox{second}{2nd}, and \colorbox{third}{3rd} best results are highlighted.}
\label{tab:bench_k562}
\setlength{\tabcolsep}{2pt}\renewcommand{\arraystretch}{1.10}\small
\begin{tabular}{l c c c c c c}
\toprule
& \multicolumn{2}{c}{Property} & \multicolumn{2}{c}{Novelty} & \multicolumn{2}{c}{Plausibility} \\
\cmidrule(lr){2-3} \cmidrule(lr){4-5} \cmidrule(lr){6-7}
Method & Pred-activity$\uparrow$ & ATAC\_acc\,$\uparrow$ & Authenticity\,$\uparrow$ & Rarity (IQR)\,$\uparrow$ & $|\Delta\,\text{Gini}|\,\downarrow$ & $|\Delta\,\text{hits}|\,\downarrow$ \\
\midrule
Train & 2.700 & 0.281 & 0.717 & - & - & - \\
\midrule
MDLM & 0.106\,$\pm$\,0.028 & 0.367\,$\pm$\,0.227 & 0.777\,$\pm$\,0.033 & $-0.03$\,$\pm$\,0.03 & \cellcolor{second}0.049\,$\pm$\,0.005 & \cellcolor{third}0.067\,$\pm$\,0.001 \\
MDLM+CG & \cellcolor{third}6.433\,$\pm$\,0.052 & 0.852\,$\pm$\,0.196 & 0.810\,$\pm$\,0.006 & $+0.41$\,$\pm$\,0.03 & 0.260\,$\pm$\,0.007 & 0.093\,$\pm$\,0.005 \\
DNA-Diff & 2.979\,$\pm$\,0.046 & 0.479\,$\pm$\,0.311 & 0.841\,$\pm$\,0.007 & $+0.13$\,$\pm$\,0.03 & \cellcolor{first}0.028\,$\pm$\,0.020 & \cellcolor{second}0.038\,$\pm$\,0.008 \\
DRAKES & \cellcolor{second}6.493\,$\pm$\,0.155 & \cellcolor{third}0.863\,$\pm$\,0.140 & \cellcolor{second}0.873\,$\pm$\,0.004 & \cellcolor{first}$+1.07$\,$\pm$\,0.08 & 0.251\,$\pm$\,0.008 & 0.069\,$\pm$\,0.011 \\
SVDD & \cellcolor{first}9.479\,$\pm$\,0.055 & \cellcolor{first}0.928\,$\pm$\,0.012 & \cellcolor{third}0.848\,$\pm$\,0.031 & \cellcolor{second}$+0.77$\,$\pm$\,0.03 & 0.305\,$\pm$\,0.002 & 0.241\,$\pm$\,0.005 \\
\midrule
LocoGen & 6.285\,$\pm$\,0.064 & \cellcolor{second}0.900\,$\pm$\,0.012 & \cellcolor{first}0.893\,$\pm$\,0.010 & \cellcolor{second}$+0.77$\,$\pm$\,0.01 & \cellcolor{third}0.094\,$\pm$\,0.003 & \cellcolor{first}0.033\,$\pm$\,0.002 \\
\bottomrule
\end{tabular}
\vspace{2pt}
\begin{minipage}{0.98\linewidth}
\raggedleft
\scriptsize
\textit{Note.} For natural sequences (Train), $\Delta\,\text{Gini} = 0.312$ and $\Delta\,\text{hits} = 0.555$
\end{minipage}
\end{table}

\begin{table}[h!]
  \centering
  \caption{Three axis benchmark in SK-N-SH cell line ($n{=}640 \times 3$ seeds, mean $\pm$ std). \colorbox{first}{1st}, \colorbox{second}{2nd}, and \colorbox{third}{3rd} best results are highlighted.}
  \label{tab:bench_sknsh}
  \setlength{\tabcolsep}{2pt}
  \renewcommand{\arraystretch}{1.10}
  \small
  \begin{tabular}{l c c c c c c}
    \toprule
    & \multicolumn{2}{c}{Property}
    & \multicolumn{2}{c}{Novelty}
    & \multicolumn{2}{c}{Plausibility} \\
    \cmidrule(lr){2-3} \cmidrule(lr){4-5} \cmidrule(lr){6-7}
    Method
    & Pred-activity $\uparrow$
    & ATAC\_acc\,$\uparrow$
    & Authenticity\,$\uparrow$
    & Rarity (IQR)\,$\uparrow$
    & $|\Delta\,\text{Gini}|\,\downarrow$
    & $|\Delta\,\text{hits}|\,\downarrow$ \\
    \midrule
    Train
    & 2.376
    & 0.212
    & 0.713
    & -
    & -
    & - \\
    \midrule
    MDLM
    & 0.112\,$\pm$\,0.026
    & 0.391\,$\pm$\,0.217
    & 0.777\,$\pm$\,0.033
    & $-0.09$\,$\pm$\,0.03
    & \cellcolor{third}0.071\,$\pm$\,0.006
    & 0.080\,$\pm$\,0.001 \\
    MDLM+CG
    & 5.047\,$\pm$\,0.101
    & \cellcolor{third}0.429\,$\pm$\,0.295
    & 0.828\,$\pm$\,0.010
    & $+0.23$\,$\pm$\,0.06
    & 0.120\,$\pm$\,0.003
    & \cellcolor{first}0.007\,$\pm$\,0.005 \\
    DNA-Diff$^{\dagger}$
    & 0.033\,$\pm$\,0.015
    & 0.079\,$\pm$\,0.087
    & 0.821\,$\pm$\,0.015
    & $+0.09$\,$\pm$\,0.03
    & \cellcolor{second}0.024\,$\pm$\,0.000
    & \cellcolor{second}0.034\,$\pm$\,0.000 \\
    DRAKES
    & \cellcolor{second}6.992\,$\pm$\,0.169
    & \cellcolor{first}0.879\,$\pm$\,0.145
    & \cellcolor{second}0.873\,$\pm$\,0.004
    & \cellcolor{second}$+1.07$\,$\pm$\,0.09
    & 0.241\,$\pm$\,0.008
    & 0.095\,$\pm$\,0.011 \\
    SVDD
    & \cellcolor{first}7.989\,$\pm$\,0.041
    & 0.356\,$\pm$\,0.025
    & \cellcolor{third}0.868\,$\pm$\,0.021
    & \cellcolor{first}$+1.25$\,$\pm$\,0.06
    & 0.260\,$\pm$\,0.001
    & 0.213\,$\pm$\,0.004 \\
    \midrule
    LocoGen
    & \cellcolor{third}5.304\,$\pm$\,0.186
    & \cellcolor{second}0.747\,$\pm$\,0.048
    & \cellcolor{first}0.897\,$\pm$\,0.016
    & \cellcolor{third}$+0.69$\,$\pm$\,0.06
    & \cellcolor{first}0.012\,$\pm$\,0.006
    & \cellcolor{third}0.081\,$\pm$\,0.004 \\
    \bottomrule
  \end{tabular}
  \vspace{2pt}
  \begin{minipage}{0.98\linewidth}
    \raggedleft
    \scriptsize
    \textit{Note.} For natural sequences (train set), $\Delta\,\text{Gini} = 0.321$ and $\Delta\,\text{hits} = 0.572$. $\dagger$ :DNA-Diff \citep{dasilva2026designing} trained on {GM12878, HepG2, hESCT0, K562}; SK-N-SH not in label set. K562-anchored seqs scored under SK-N-SH oracle head.
  \end{minipage}
\end{table}

\subsection{Masked separator faithfully disentangles invariant and environmental subsequences.}
\label{app:effect_of_masking}
\begin{figure}[h!]
\centering
\includegraphics[width=0.5\linewidth]{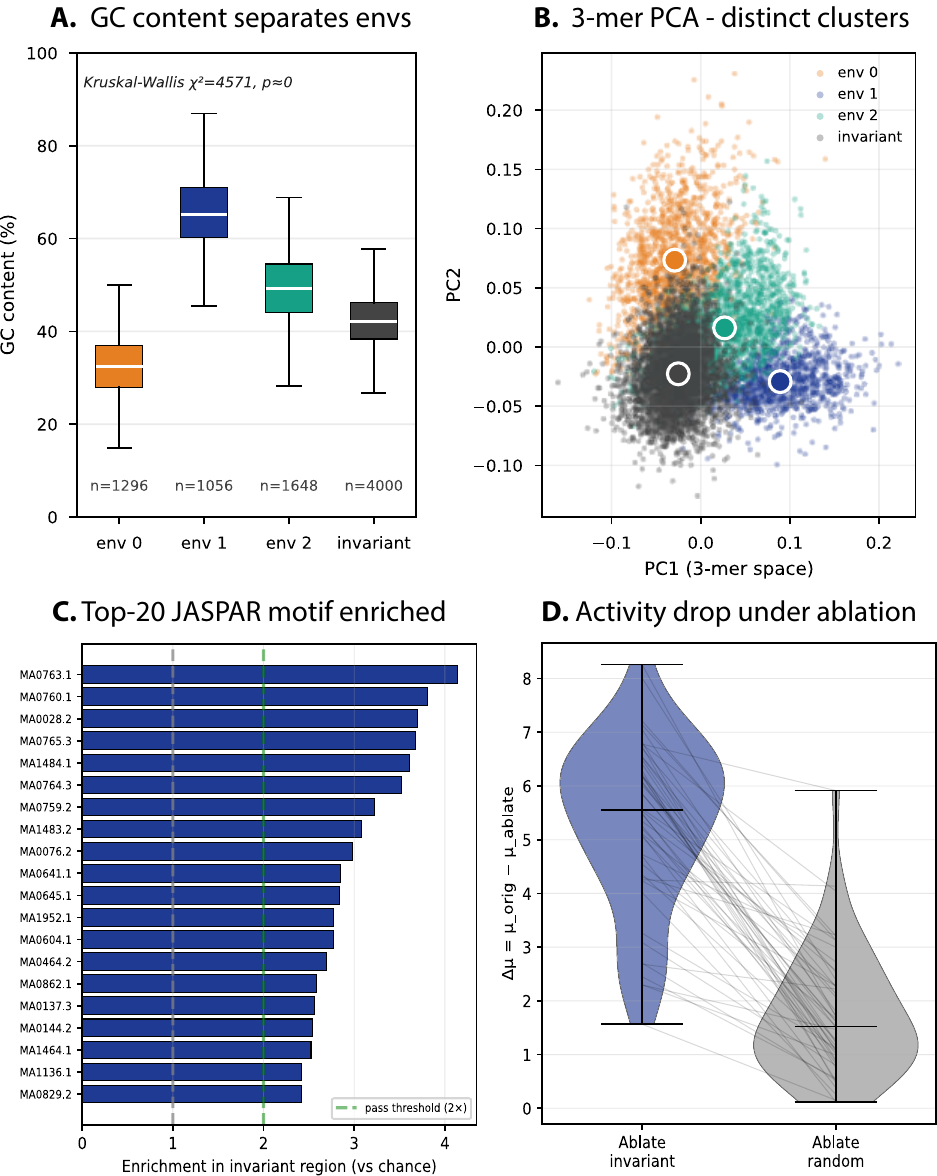}
\caption{Sequence-level invariance learning recovers regulatory grammar.}
\label{fig:separator_diagnostic}
\end{figure}

We next asked whether the masked separator module recovers a meaningful invariant/environmental decomposition. Three lines of evidence support that it does (Fig.~\ref{fig:separator_diagnostic}). First, the three environmental subsequences exhibit clearly distinct GC-content distributions while the invariant subsequence sits near the natural baseline (Fig.~\ref{fig:separator_diagnostic}A), and the same separation holds in 3-mer PCA space, where each environment forms a distinct cluster around its own centroid (Fig.~\ref{fig:separator_diagnostic}B). Second, the invariant subsequence is enriched for canonical activity-driving JASPAR motifs, predominantly ETS-family factors (Fig.~\ref{fig:separator_diagnostic}C). Third, ablating only the invariant subsequence causes a substantially larger drop in oracle-predicted activity than ablating an equally-sized random region (Fig.~\ref{fig:separator_diagnostic}D), showing that the invariant region is functionally responsible for the predicted activity rather than being an arbitrary partition.

\subsection{Ablation study}

\begin{figure}[h!]
    \centering
    \includegraphics[width=0.99\linewidth]{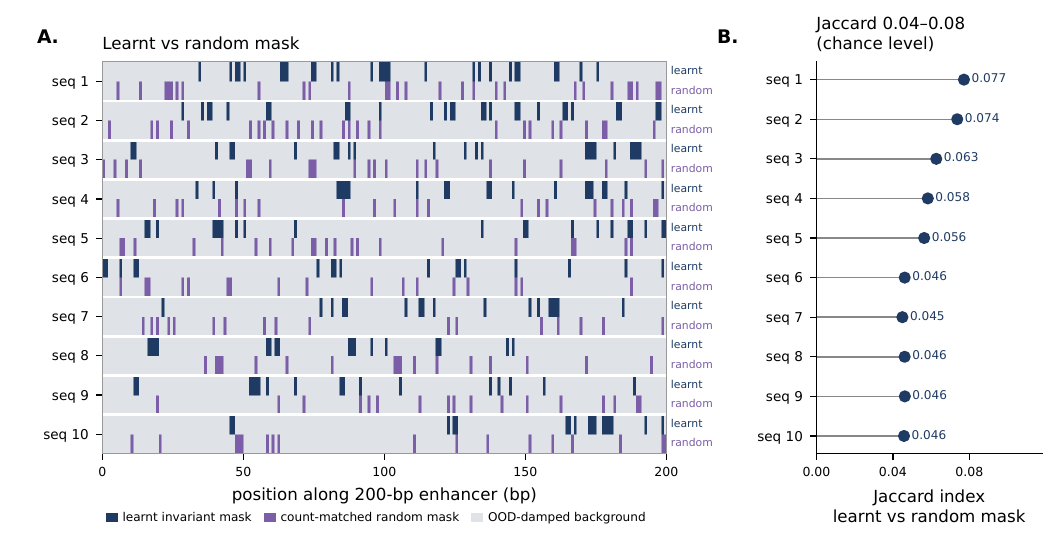}
    \vspace{5pt}
    \caption{\textbf{The learnt invariant region is not a random subset of positions.} (A) For ten HepG2 enhancers, the learnt invariant mask (dark) and a count-matched random mask (purple) along the 200bp window; the light band is the OOD-damped background. Both masks select the same number of positions, so any difference in the ablation is attributable to which positions are spared rather than how many. (B) Per-sequence Jaccard index between the two masks.}
    \label{fig:learnt_vs_random}
\end{figure}

To better interpret the global-OOD setting in \S\ref{sec:ablation}, we conducted an additional experiment
in which the same number of masked positions was distributed
uniformly at random across the sequence
(Fig.~\ref{fig:learnt_vs_random}).
We confirmed that these randomly selected positions
exhibited minimal overlap with the invariant region.

\subsubsection{Effect of OOD control strength}
\label{app:lambda_sweep}
Table~\ref{tab:lambda_sweep} studies the effect of the OOD control strength $\gamma$. 
Without OOD control the model achieves reasonable predicted activity but shows weak ATAC accessibility, low authenticity, limited rarity, and a large deviation in $|\Delta\text{Gini}|$. 
Increasing $\gamma$ to 0.3 substantially improves ATAC accessibility, authenticity, rarity, and both plausibility metrics, indicating that moderate OOD control leads to generate more plausible sequences. 
At $\gamma=0.7$, \textsc{LocoGen} achieves the best authenticity, rarity, and $|\Delta\text{Gini}|$, while also maintaining strong predicted activity and the highest ATAC accessibility. 
Although $\gamma=0.3$ gives slightly higher \textit{pred-activity} and the lowest $|\Delta\text{hits}|$, the $\gamma=0.7$ setting provides the most balanced operating point across property, novelty, and plausibility. 
These results support the role of OOD control as a calibration mechanism that controls the trade-off between aggressive property optimization and distributional realism.

\newcommand{\best}[1]{\textbf{\boldmath #1}}

\begin{table}[h!]
\centering
\caption{OOD strength sweep on HepG2.}
\label{tab:lambda_sweep}
\small
\setlength{\tabcolsep}{4pt}
\begin{tabular}{c c c c c c c}
\toprule
$\gamma$ & $\textrm{Pred-activity}\uparrow$ & ATAC$_{acc}\uparrow$ & Auth.$\uparrow$ & Rarity (IQR)$\uparrow$ & $|\Delta\text{Gini}|\downarrow$ & $|\Delta\text{hits}|\downarrow$ \\
\midrule
$0.0$ & $4.516 \pm 0.151$ & $0.481 \pm 0.022$ & $0.809 \pm 0.011$ & $+0.22 \pm 0.01$ & $0.169 \pm 0.010$ & $0.043 \pm 0.005$ \\
$0.3$ & $4.738 \pm 0.061$ & $0.651 \pm 0.007$ & $0.870 \pm 0.004$ & $+0.53 \pm 0.03$ & $0.095 \pm 0.003$ & $0.027 \pm 0.002$ \\
$0.7$ & $4.679 \pm 0.0$ & $0.716 \pm 0.019$ & $0.906 \pm 0.005$ & $+0.71 \pm 0.00$ & $0.055 \pm 0.004$ & $0.057 \pm 0.002$ \\
\bottomrule
\end{tabular}
\end{table}                             

\subsubsection{Segment and shuffle baseline}
\label{app:seg_and_shuffle}
We uniformly sampled regulatory sequences from the training set, preserved positions designated as foreground, and shuffled the remaining background positions. Here, foreground and background refer to regions selected for preservation and modification, respectively, rather than the invariant and environment components defined in \textsc{LocoGen}. We evaluated combinations of three foreground segmentation methods $\times$ two background replacement strategies.
\begin{itemize}
    \item \textbf{Foreground segmentation}
    \begin{itemize}
        \item Position-specific scoring matrix (PSSM) motif scan
        \item GC content-based segmentation
        \item Predictor saliency
    \end{itemize}

    \item \textbf{Background replacement}
    \begin{itemize}
        \item Uniform random replacement
        \item Altschul-Erikson dinucleotide shuffle (i.e., dinucleotide-preserving) ~\citep{altschul1985significance}
    \end{itemize}
\end{itemize}

\begin{table}[h!]
\centering
\caption{Comparison with non-generative segment-and-shuffle controls on HepG2.}
\label{tab:segment_shuffle}
\resizebox{\textwidth}{!}{%
\begin{tabular}{llcccccc}
\toprule
\multirow{2}{*}{Segmentation} &
\multirow{2}{*}{Replacement} &
\multicolumn{2}{c}{Property} &
\multicolumn{2}{c}{Novelty} &
\multicolumn{2}{c}{Plausibility} \\
\cmidrule(lr){3-4}
\cmidrule(lr){5-6}
\cmidrule(lr){7-8}
& & Pred-act. $\uparrow$
  & ATAC$_{\mathrm{acc}}$ $\uparrow$
  & Authent. $\uparrow$
  & Rarity (IQR) $\uparrow$
  & $|\Delta \mathrm{Gini}|$ $\downarrow$
  & $|\Delta \mathrm{hits}|$ $\downarrow$ \\
\midrule
JASPAR PSSM & Uniform random
  & 0.275 & 0.206 & 0.928 & $+0.25$ & 0.155 & 0.114 \\
JASPAR PSSM & Dinuc.-preserving
  & 0.122 & 0.092 & 0.815 & $-0.12$ & 0.069 & 0.066 \\
\midrule
GC content & Uniform random
  & 0.359 & 0.252 & 0.907 & $+0.13$ & 0.148 & 0.129 \\
GC content & Dinuc.-preserving
  & 0.078 & 0.086 & 0.809 & $-0.16$ & 0.054 & 0.076 \\
\midrule
Saliency & Uniform random
  & 0.288 & 0.249 & 0.961 & $+0.30$ & 0.167 & 0.138 \\
Saliency & Dinuc.-preserving
  & 0.199 & 0.129 & 0.840 & $-0.03$ & 0.101 & 0.095 \\
\midrule
LearnedMask & Uniform random
  & 0.442 & 0.280 & 0.955 & $+0.38$ & 0.166 & 0.171 \\
LearnedMask & Dinuc.-preserving
  & 0.287 & 0.109 & 0.823 & $-0.05$ & 0.067 & 0.113 \\
\midrule
\multicolumn{2}{c}{LocoGen}
  & 4.679 & 0.716 & 0.906 & $+0.71$ & 0.055 & 0.057 \\
\bottomrule
\end{tabular}%
}
\end{table}

\subsection{Robust to choice of oracle}
\label{app:robust_oracle}
To demonstrate the generalizability of \textsc{LocoGen} across different experimental settings, we evaluated robustness to different oracles for activity and accessibility evaluation.
We evaluated our method using two publicly available alternative oracles released by prior work \citep{li2024derivative, genentech_human_atac_catlas_model}.
\begin{itemize}
    \item Activity oracle \citep{li2024derivative}: an independent Enformer-based predictor trained on the same Gosai dataset, but with a different train/test split and substantially more training data than the DRAKES oracle.
    \item Accessibility oracle \citep{genentech_human_atac_catlas_model}: an independent Enformer-based predictor trained on the Human ATAC Atlas \citep{zhang2021single}, a different dataset for chromatin accessibility prediction.
    
\end{itemize}
Overall, \textsc{LocoGen} consistently maintains strong property performance across independent evaluation oracles, demonstrating that our conclusions are not tied to a specific oracle. 

\begin{table}[h!]
\centering
\small
\setlength{\tabcolsep}{5pt}
\caption{\textbf{Robustness to oracle choice (HepG2).} Beyond the DRAKES evaluation
oracles used in the main text, we add two predictors that differ in training split and
in data source, respectively. Rank across the six methods in parentheses; values are
mean $\pm$ s.d.\ where seed replicates are available.}
\begin{tabular}{lcccc}
\toprule
& \multicolumn{2}{c}{Activity} & \multicolumn{2}{c}{Accessibility} \\
\cmidrule(lr){2-3}\cmidrule(lr){4-5}
Method
& \makecell{DRAKES\\Pred-activity}
& \makecell{SVDD DNA\_evaluation\\\emph{alternative training split}}
& \makecell{DRAKES\\ATAC\_acc}
& \makecell{human-atac-catlas\\\emph{different data source}} \\
\midrule
MDLM            & 0.157 (6)          & $0.408 \pm 0.013$ (6)          & 0.329 (6)          & $0.050 \pm 0.009$ (6) \\
MDLM+CG         & 4.516 (4)          & $4.434 \pm 0.097$ (4)          & 0.481 (4)          & $0.459 \pm 0.014$ (5) \\
DNA-Diff        & 2.595 (5)          & $2.721 \pm 0.046$ (5)          & 0.688 (3)          & $0.517 \pm 0.020$ (4) \\
DRAKES          & 5.610 (2)          & $5.135 \pm 0.076$ (2)          & 0.804 (1)          & $0.829 \pm 0.006$ (1) \\
SVDD            & 6.830 (1)          & $6.187 \pm 0.018$ (1)          & 0.442 (5)          & $0.547 \pm 0.010$ (3) \\
\textbf{LocoGen}& \textbf{4.679} (3) & $\mathbf{4.660 \pm 0.011}$ (3) & \textbf{0.716} (2) & $\mathbf{0.729 \pm 0.002}$ (2) \\
\bottomrule
\end{tabular}
\label{tab:robustness-hepg2}
\end{table}

\begin{table}[h!]
\centering
\small
\setlength{\tabcolsep}{5pt}
\caption{\textbf{Robustness to oracle choice (K562).} Same four oracles and conventions
as Table~\ref{tab:robustness-hepg2}; rank across the six methods in parentheses, values
are mean $\pm$ s.d.\ where seed replicates are available.}
\begin{tabular}{lcccc}
\toprule
& \multicolumn{2}{c}{Activity} & \multicolumn{2}{c}{Accessibility} \\
\cmidrule(lr){2-3}\cmidrule(lr){4-5}
Method
& \makecell{DRAKES\\Pred-activity}
& \makecell{SVDD DNA\_evaluation\\\emph{alternative training split}}
& \makecell{DRAKES\\ATAC\_acc}
& \makecell{human-atac-catlas\\\emph{different data source}} \\
\midrule
MDLM            & 0.106 (6)          & $0.381 \pm 0.031$ (6)          & 0.367 (6)          & $0.055 \pm 0.006$ (6) \\
MDLM+CG         & 6.433 (3)          & $6.191 \pm 0.042$ (2)          & 0.852 (4)          & $0.688 \pm 0.022$ (4) \\
DNA-Diff        & 2.979 (5)          & $3.038 \pm 0.036$ (5)          & 0.479 (5)          & $0.359 \pm 0.009$ (5) \\
DRAKES          & 6.493 (2)          & $6.031 \pm 0.114$ (4)          & 0.863 (3)          & $0.837 \pm 0.005$ (3) \\
SVDD            & 9.479 (1)          & $8.490 \pm 0.064$ (1)          & 0.928 (1)          & $0.899 \pm 0.005$ (1) \\
\textbf{LocoGen}& \textbf{6.285} (4) & $\mathbf{6.085 \pm 0.014}$ (3) & \textbf{0.900} (2) & $\mathbf{0.846 \pm 0.002}$ (2) \\
\bottomrule
\end{tabular}
\label{tab:robustness-k562}
\end{table}

\begin{table}[h!]
\centering
\small
\setlength{\tabcolsep}{5pt}
\caption{\textbf{Robustness to oracle choice (SK-N-SH).} Same four oracles and
conventions as Table~\ref{tab:robustness-hepg2}; rank across the six methods in
parentheses, values are mean $\pm$ s.d.\ where seed replicates are available.}
\begin{tabular}{lcccc}
\toprule
& \multicolumn{2}{c}{Activity} & \multicolumn{2}{c}{Accessibility} \\
\cmidrule(lr){2-3}\cmidrule(lr){4-5}
Method
& \makecell{DRAKES\\Pred-activity}
& \makecell{SVDD DNA\_evaluation\\\emph{alternative training split}}
& \makecell{DRAKES\\ATAC\_acc}
& \makecell{human-atac-catlas\\\emph{different data source}} \\
\midrule
MDLM            & 0.112 (5)          & $0.340 \pm 0.033$ (6)          & 0.391 (4)          & $0.030 \pm 0.005$ (6) \\
MDLM+CG         & 5.047 (4)          & $4.970 \pm 0.104$ (4)          & 0.429 (3)          & $0.377 \pm 0.020$ (3) \\
DNA-Diff        & 0.033 (6)          & $0.720 \pm 0.021$ (5)          & 0.079 (6)          & $0.040 \pm 0.002$ (5) \\
DRAKES          & 6.992 (2)          & $6.132 \pm 0.153$ (2)          & 0.879 (1)          & $0.839 \pm 0.007$ (1) \\
SVDD            & 7.989 (1)          & $6.714 \pm 0.103$ (1)          & 0.356 (5)          & $0.192 \pm 0.004$ (4) \\
\textbf{LocoGen}& \textbf{5.304} (3) & $\mathbf{5.294 \pm 0.156}$ (3) & \textbf{0.747} (2) & $\mathbf{0.629 \pm 0.020}$ (2) \\
\bottomrule
\end{tabular}
\label{tab:robustness-sknsh}
\end{table}

\subsection{Robust to hyperparameter sweep}
\label{app:robust_hyperparam}
We ran a hyperparameter sweep of both our method and the baselines, sweeping each over the range specified in its paper or official repository. Across these sweeps, \textsc{LOCOGEN} retains its balanced three-axis advantage, confirming that the result does not rely on a cherry-picked configuration.

\begin{figure}[h!]
    \centering
    \includegraphics[width=0.99\linewidth]{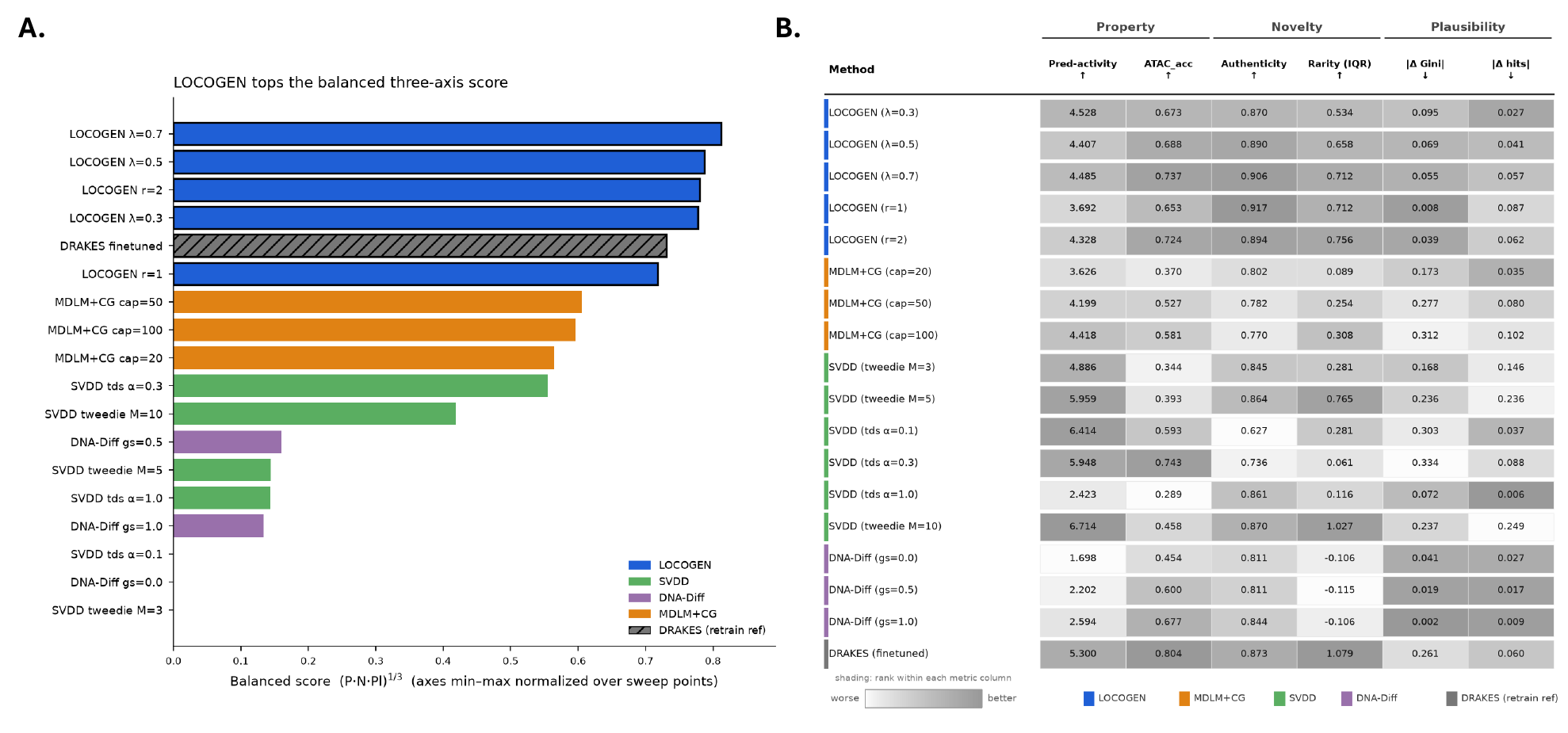}
    \caption{(A). Balanced score over three evaluation axis where \textsc{LocoGen} achieves the highest balanced score across sweep points. (B) Evaluation metrics across the sweep, with one row per hyperparameter setting. The value in parentheses indicates each method's strength setting:
$\gamma$ denotes OOD/calibration strength (\texttt{ood\_lambda});
$r$ denotes the property-pull ratio (\texttt{alpha\_ratio}) at $\gamma=0.7$
(the paper default $r=3.0$ corresponds to the $\gamma=0.7$ row,
while $r=1,2$ are lower-pull variants);
$\mathrm{cap}$ denotes the MDLM+CG classifier-guidance $\alpha$-cap
(\texttt{alpha\_cap});
$M$ denotes the number of SVDD candidates per step (Tweedie/SVDD-PM);
$\alpha_{\mathrm{TDS}}$ denotes the SVDD-TDS importance-resampling temperature;
and $\mathrm{gs}$ denotes the DNA-Diff classifier-free guidance scale
(\texttt{cond\_weight}; author default $1.0$).
DRAKES fine-tunes model weights for a fixed reward rather than steering
generation at inference time, and is therefore included as a retraining
reference rather than a directly comparable operating point.}
    \label{fig:hyperparameter_sweep}
\end{figure}
\section{Discussion}
\label{app:discussion}

\subsection{Limitations.}
Our experiments are conducted on a single backbone, MDLM, to isolate the effect of localized OOD guidance from backbone-dependent variation. Validation on alternative backbones, such as flow matching or autoregressive models, and a full comparison across backbones are left to future work. Computational evaluation alone, while standard in this domain, cannot substitute for \emph{in vivo} experimental validation of generated sequences, a precondition for treating any sampling-time guidance method as practically useful. 

\subsection{Social Impact}
Our work may contribute to bioengineering by improving the ability to generate novel biological sequences beyond the training distribution. Such capability can expand the searchable design space and accelerate the discovery of sequences with useful properties. In practical settings, this could reduce the cost and time required for experimental screening, while supporting applications in therapeutic development, enzyme engineering, antibody design, and synthetic biology.

However, the broader impact of biological sequence generation depends on how the technology is used. Since generative models can produce sequences that differ substantially from naturally observed examples, they may also introduce biosafety and dual-use risks. For example, the same methods that help design beneficial proteins or regulatory elements could potentially be misused to explore harmful biological functions. Therefore, responsible use requires careful consideration of dataset provenance, model access, downstream experimental validation, and risk assessment tailored to the intended application.

\clearpage
\newpage

\end{document}